%% file: main.tex
\documentclass{article} %
\usepackage{iclr2027_conference,times}
\newif\ifarxiv
\arxivtrue                 %
\ifarxiv\iclrfinalcopy\fi

\input{math_commands.tex}

\usepackage{amsmath,amssymb}
\usepackage{hyperref}
\usepackage{url}
\usepackage{graphicx}
\usepackage{booktabs}
\usepackage{multirow}
\usepackage[table]{xcolor}
\usepackage{pifont}
\usepackage{subcaption}
\usepackage{makecell}
\usepackage{enumitem}
\usepackage{float}
\usepackage[section]{placeins}
\usepackage{xspace}
\usepackage[capitalize,nameinlink]{cleveref}
\crefname{section}{Section}{Sections}
\Crefname{section}{Section}{Sections}
\crefname{figure}{Figure}{Figures}
\crefname{table}{Table}{Tables}
\crefname{equation}{Eq.}{Eqs.}
\crefname{appendix}{Appendix}{Appendices}
\usepackage{wrapfig}
\AtBeginDocument{\setlength{\abovedisplayskip}{5pt plus 2pt minus 2pt}\setlength{\belowdisplayskip}{5pt plus 2pt minus 2pt}}
\makeatletter
\def\section{\@startsection {section}{1}{\z@}{-1.3ex plus -0.4ex minus -.2ex}{0.8ex plus 0.2ex minus 0.1ex}{\large\sc\raggedright}}
\def\subsection{\@startsection{subsection}{2}{\z@}{-1.0ex plus -0.3ex minus -.2ex}{0.5ex plus .2ex}{\normalsize\sc\raggedright}}
\def\paragraph{\@startsection{paragraph}{4}{\z@}{0.8ex plus 0.3ex minus .2ex}{-1em}{\normalsize\bf}}
\makeatother

\newcommand{\best}[1]{{\color{red}#1}}   %
\newcommand{\cmark}{\ding{51}}
\newcommand{\xmark}{\ding{55}}

\newcommand{\ie}{\emph{i.e}.,\xspace}

\title{3AM: 3egment Anything with Adaptive\\ Geometric Consistency in Videos}

\author{Yang-Che Sun$^{1}$, Cheng Sun$^{2}$, Chin-Yang Lin$^{1}$, Fu-En Yang$^{2}$, \\
\textbf{Min-Hung Chen$^{2}$, Yen-Yu Lin$^{1}$, Yu-Lun Liu$^{1}$} \\
$^{1}$Department of Computer Science, National Yang Ming Chiao Tung University \quad $^{2}$NVIDIA Research \\
\texttt{jkarly.md11@nycu.edu.tw, yulunliu@cs.nycu.edu.tw}
}

\begin{document}

\maketitle
\ifarxiv\lhead{Preprint. Under review.}\fi

\input{00_abstract}

\begin{figure}[H]
    \centering
    \includegraphics[width=\textwidth]{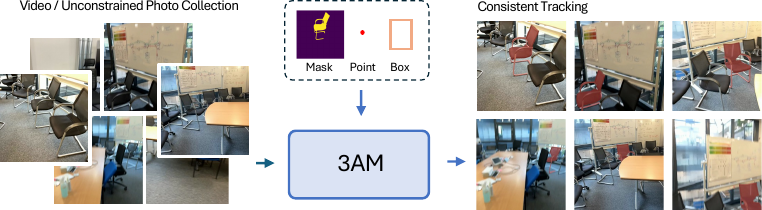}\\[2mm]
    \includegraphics[width=\textwidth]{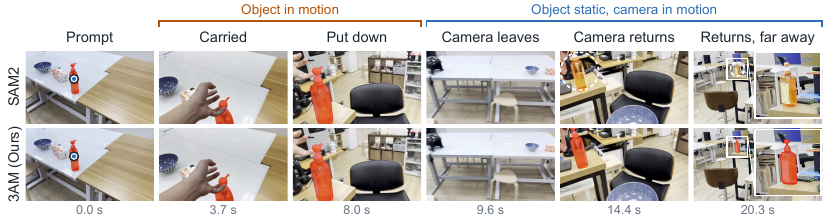}
    \caption{\textbf{Overview of 3AM and tracking through object motion and reappearance.}
    Upper: given a video or unordered photo collection and a mask, point, or box prompt on one frame, 3AM tracks the object across all views, online and from RGB only, adapting between geometry and appearance.
    Lower: in a phone video prompted with one click, the bottle is carried, put down elsewhere, and revisited twice; both trackers follow it while it moves, but only 3AM re-identifies it on both returns, whereas SAM2~\citep{ravi2024sam2} never recovers it (dashed circle, magnified in the inset).}
    \label{fig:overview}\label{fig:teaser}\label{fig:in-the-wild}
\end{figure}

\input{01_intro}
\input{02_related}

\input{03_method}
\input{04_experiments}
\input{05_conclusion}

\input{06_statements}
\bibliography{main}
\bibliographystyle{iclr2027_conference}

\clearpage
\appendix
\input{10_appendix}

\end{document}

%% file: math_commands.tex
\usepackage{amsmath,amsfonts,bm}

\def\eqref#1{equation~\ref{#1}}

\def\1{\bm{1}}

\DeclareMathAlphabet{\mathsfit}{\encodingdefault}{\sfdefault}{m}{sl}
\SetMathAlphabet{\mathsfit}{bold}{\encodingdefault}{\sfdefault}{bx}{n}

%% file: 00_abstract.tex
\vspace{-3mm}
\begin{abstract}
\vspace{-3mm}
Consistent video object tracking must preserve identity under large viewpoint changes, temporary absence, and nearby distractors.
Memory-based trackers such as SAM2 rely on appearance matching and fail in these cases, whereas 3D-based methods preserve identity across views but need camera poses, depth, or a reconstructed point cloud, run offline, and are not promptable.
We present 3AM, a promptable online tracker that achieves geometric consistency from RGB alone by coupling memory-based tracking with the online multi-view reconstruction model MUSt3R.
We identify the MUSt3R layers whose features preserve instance identity across views and fuse them into the tracker's image features by feature-wise modulation whose parameters are decoded from \emph{routing tokens}.
Geometry alone is insufficient, since a location cue holds only while the object stays in place; the routing tokens therefore traverse the tracker's memory attention and decide per frame how much the fusion should rely on geometry, so that one model handles static and moving objects without a hand-designed fallback.
On wide-baseline benchmarks from ScanNet++ and Replica, 3AM reaches 90.6\% IoU and 65.7\% tracking recall on our selected subset of ScanNet++ (+12.3 and +28.2 points over the best prior tracker), matches or exceeds SAM2 on standard VOS and VOT benchmarks, and transfers zero-shot to ego-exo correspondence on Ego-Exo4D.
Project page: \url{https://jayisaking.github.io/3AM-Page/}
\end{abstract}

%% file: 01_intro.tex
\section{Introduction}
\label{sec:intro}

\begin{figure}[t]
    \centering
    \includegraphics[width=\textwidth]{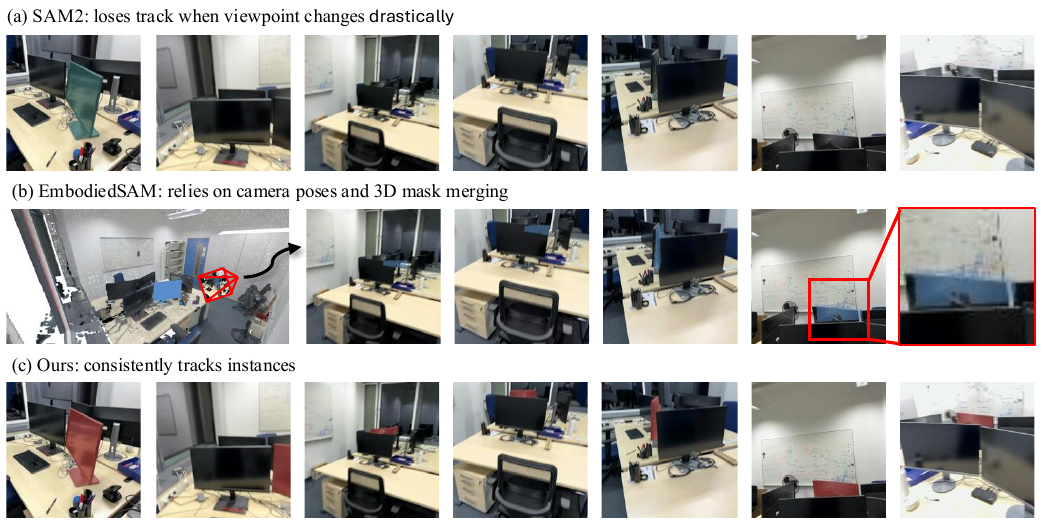}
    \caption{\textbf{Two existing paradigms and ours.}
    (a) Memory-based trackers such as SAM2~\citep{ravi2024sam2} lose the object under a large viewpoint change.
    (b) 3D segmentation methods preserve identity in scene coordinates but need camera poses and 3D mask merging, and propagate reconstruction errors.
    (c) 3AM preserves identity across drastic viewpoint changes from RGB alone.}
    \label{fig:intro-comparison}
\end{figure}

Video object segmentation (VOS) propagates a prompt on one frame to masks of the same object in all other frames, and for embodied agents the central requirement is consistency.
As a robot or wearable camera moves through a room, the object leaves the view and reappears from another direction, at another distance, or next to similar objects (\cref{fig:in-the-wild}, lower); the tracker must return the same object, and no mask while the object is absent.
Memory-based models such as SAM2~\citep{ravi2024sam2} match each frame against stored appearance, which fails under large viewpoint changes because no stored view matches (\cref{fig:intro-comparison}a); on ScanNet++~\citep{yeshwanth2023scannet++}, SAM2 and its memory-selection extensions~\citep{ding2024sam2long,videnovic2025distractor} degrade sharply, as memory management cannot recover an object whose appearance has changed.

Appearance features lack the object's location.
3D-based methods preserve identity by construction, by segmenting point clouds~\citep{schult2022mask3d,kolodiazhnyi2024oneformer3d} or by lifting and merging 2D masks~\citep{yang2023sam3d,nguyen2024open3dis,zhao2025sam2object}, but need poses, depth, and a reconstruction whose cost grows with sequence length, and are neither promptable nor online (\cref{fig:intro-comparison}b).
Trackers thus offer the interface without location, and 3D-based methods the reverse.

We bridge this gap by coupling a promptable memory-based tracker with an online reconstruction model that provides geometry at every frame (\cref{fig:intro-comparison}c).
MUSt3R~\citep{cabon2025must3r} predicts point maps in a common frame from RGB, online and without poses; we identify its layers that preserve instance identity across views and fuse them into SAM2's image features by lightweight feature-wise modulation, so that memory attention, mask decoder, and memory encoder see geometry-aware features.
Training uses 2D masks only, and inference needs no poses or depth.

Integrating geometry is necessary but not sufficient.
Geometry is reliable only while the object stays in place, but objects are picked up, carried, and put down, so a tracker that always trusts geometry fails on a carried object and one that always trusts appearance fails on reappearance.
We make this trade-off adaptive with \emph{routing tokens} that traverse the tracker's memory attention, read the memory bank and the current frame, and decide how much the next frame's fusion relies on each geometric source; the geometric path opens for static objects and closes for moving ones.
In the lower part of \cref{fig:in-the-wild}, one model follows the carried bottle and re-identifies it once put down.

Our contributions are threefold.
\textbf{(i) A view-consistency gap in video object tracking.}
Memory-based trackers lose identity under large viewpoint changes and reappearance, and 3D-based methods need poses, depth, and offline reconstruction; our ScanNet++ and Replica protocol isolates wide-baseline reappearance.
\textbf{(ii) Geometry-consistent tracking from RGB alone.}
A layer-wise analysis of MUSt3R identifies the layers that preserve instance identity across views, which lightweight feature-wise modulation fuses into a promptable memory-based tracker.
\textbf{(iii) Adaptive routing between geometry and appearance.}
Routing tokens in SAM2's memory attention set every frame's fusion parameters, so one model tracks static and moving objects without a hand-designed fallback.
3AM raises IoU on our ScanNet++ Selected Subset from 78.3 (SAM2Long) to 90.6 and Tracking Recall from 37.5 (DAM4SAM) to 65.7, matches or exceeds SAM2 on dynamic benchmarks, and transfers zero-shot to ego-exo correspondence on Ego-Exo4D~\citep{grauman2024egoexo4d}.

%% file: 02_related.tex
\section{Related Work}
\label{sec:related}

\paragraph{Video Object Segmentation.}
VOS propagates a mask from one frame through a video~\citep{perazzi2016benchmark,caelles2017one,voigtlaender2019feelvos,maninis2018video,yang2019video,oh2019video,bhat2020learning}.
Memory-based methods match past frames by spatio-temporal attention~\citep{yang2021associating,cheng2022xmem,cheng2021rethinking,liu2022learning,park2022per,xu2022reliable,yang2024scalable,yang2022decoupling,li2024univs,cheng2024putting}, with later work on memory size~\citep{zhou2024rmem,liu2025livos}, unified transformers~\citep{li2024onevos,zhang2025jointformer}, anchor queries~\citep{zhou2024improving}, and decoupled segmentation and propagation~\citep{cheng2023tracking}.
SAM2~\citep{ravi2024sam2} adds streaming memory to SAM~\citep{kirillov2023segment}; follow-ups improve memory selection~\citep{ding2024sam2long,videnovic2025distractor,yang2024samurai,yang2025mosam}, efficiency~\citep{zhou2025edgetam}, and language grounding~\citep{cuttano2025samwise,yuan2025sa2va}, and SAM~3~\citep{carion2025sam3} adds concept prompts.
Recent variants separate short- and long-term memory~\citep{kim2026sam2dual,alansari2025rethinking}, gate memory writes by presence~\citep{chen2026streamdam}, unify mask, box, and point granularities~\citep{zhang2025sam2pp}, add motion and semantic cues~\citep{zhu2026samosa}, or build object concepts with a vision-language model~\citep{zhang2025advancing}.
Long-term benchmarks require absence reporting and re-detection~\citep{valmadre2018long,tang2023egotracks}, object permanence is learned for multi-object tracking~\citep{tokmakov2021learning}, appearance memories re-identify indoor re-entries~\citep{diaz2026remind}, and Whareformer~\citep{chalk2026whareformer} associates observations of moved objects with tracks through appearance and 3D-location memories, given masks from a separate segmenter and depth aligned to a reconstruction.
All rely on appearance matching, which benchmarks with large viewpoint changes and reappearance~\citep{ding2025mosev2,hong2023lvos,zhang2025advancing} show to be fragile; 3AM instead makes a promptable tracker's memory geometry-aware.

\paragraph{Consistency through 3D Segmentation.}
Methods that preserve identity across views operate in scene coordinates.
Proposal-based methods segment point clouds~\citep{schult2022mask3d,jiang2020pointgroup,vu2022softgroup,sun2023superpoint,kolodiazhnyi2024oneformer3d,takmaz2023openmask3d}, and lifting methods merge per-view masks in 3D by linear assignment, contrastive fusion, or view-consensus clustering~\citep{siddiqui2023panoptic,bhalgat2023contrastive,yan2024maskclustering,yang2023sam3d,nguyen2024open3dis,yin2024sai3d,jung2025details,zhao2025sam2object}, sometimes through feature fields or Gaussians~\citep{kerr2023lerf,qin2024langsplat,ye2024gaussian,shen2024flashsplat}, as do object-centric maps for embodied agents~\citep{gu2024conceptgraphs}; SAM3D~\citep{yang2023sam3d} is unrelated to the single-image SAM~3D models~\citep{chen2026sam,yang2026sam}.
Most need posed RGB-D and offline processing; online variants work on posed RGB-D~\citep{xu2025embodiedsam,wang2026online} or monocular RGB with reconstruction priors~\citep{du2026moonseg3r,gelencser2026scenevggt} but output scene-level instances, not a prompted object.
Merging is needed because per-view masks are inconsistent~\citep{siddiqui2023panoptic,bhalgat2023contrastive,peng2023openscene,jia2024sceneverse}; 3AM instead makes the 2D tracks consistent, so their projection into 3D needs little merging (\cref{sec:exp-3d}).

\paragraph{Feed-Forward 3D Reconstruction and 3D-Aware Segmentation.}
DUSt3R-style models regress point maps from RGB~\citep{wang2024dust3r,leroy2024grounding} and have become multi-view and online.
Spann3R~\citep{wang20253d} adds a spatial memory, MUSt3R~\citep{cabon2025must3r} and CUT3R~\citep{wang2025continuous} keep a persistent memory or state, STream3R~\citep{lan2025stream3r} and StreamVGGT~\citep{zhuo2026streaming} use causal attention, Point3R~\citep{wu2025point3r}, LingBot-Map~\citep{chen2026lingbotmap}, and Anchor3R~\citep{tao2026anchor3r} bound the memory of long streams, MonST3R~\citep{zhang2024monst3r} and Easi3R~\citep{chen2025easi3r} handle dynamic scenes, and SLAM systems build on these priors~\citep{murai2025mast3r,liu2025slam3r,maggio2026vggt}, whereas VGGT~\citep{wang2025vggt}, $\pi^3$~\citep{wang2025pi3}, Depth Anything~3~\citep{lin2026depthanything3}, Fast3R~\citep{yang2025fast3r}, and MapAnything~\citep{keetha2026mapanything} process frame sets offline.
For segmentation, PanSt3R~\citep{zust2025panst3r} predicts panoptic labels from MUSt3R features offline, SegMASt3R~\citep{Jayanti2025SegMASt3R} matches segments between two images, ODIN~\citep{jain2024odin} fuses 2D and 3D backbones, IGGT~\citep{li2025iggt} clusters learned 3D-consistent instance embeddings, and SAM-V~\citep{gong2026samv} and MV-SAM~\citep{jeong2026mv} inject VGGT features or point maps into SAM to segment one object in a fixed set of static views without temporal memory.
None is both promptable and online.
Probing studies measure the 3D awareness of 2D foundation features~\citep{el2024probing,man2024lexicon3d,chen2025feat2gs}, find epipolar geometry in the middle layers of reconstruction models~\citep{bratulic2025geometric}, and show that VGGT layers specialize~\citep{ziliotto2026vggt}; \cref{sec:feature-analysis} measures instance identity across MUSt3R layers.

\paragraph{Tracking through 4D Reconstruction.}
Point trackers follow sparse points in 2D~\citep{doersch2023tapir,karaev2024cotracker}, online with a memory~\citep{aydemir2025track,aydemir2026track}, or in 3D~\citep{xiao2024spatialtracker,ngo2025delta}, and long-range tracks with semantic features can prompt SAM2 to segment moving objects~\citep{Huang_2025_seganymotion}.
Among feed-forward 4D models, TAPIP3D~\citep{zhang2025tapip3d} and SpatialTrackerV2~\citep{xiao2025spatialtrackerv2} track points in a camera-stabilized 3D space, TrackingWorld~\citep{lu2025trackingworld} optimizes world-centric trajectories per video, D4RT~\citep{zhang2026d4rt}, 4RC~\citep{luo2026fourrc}, and Track4World~\citep{lu2026track4world} encode a video once and answer motion queries for any pixel and time, and Point4D~\citep{jeon2026point4d} extends such queries to multi-hundred-frame videos by chunking.
They process whole videos or long chunks offline and output trajectories rather than masks, so identity after a long absence still rests on a mask model; we pair their tracks with SAM2 as baselines.
SAM4D~\citep{xu2025sam4d} extends SAM2's memory to camera and LiDAR streams and GOT-Edit~\citep{chen2026got} injects VGGT features into a box tracker, but the former needs LiDAR and the latter outputs boxes, and IGGT4D~\citep{zou2026iggt4d} streams instance-grounded geometry but clusters all instances of a scene rather than following a prompt and needs 4D instance annotations.
Closest to 3AM, G$^2$TAM~\citep{zhu2026g2tam} builds on $\pi^3$ and uses aligned geometric representations as an implicit memory for promptable tracking, but processes image sets offline; 3AM keeps an online tracker's appearance memory and learns per frame how much to rely on geometry, so that one model also tracks moving objects.

\paragraph{Cross-View Object Correspondence.}
Ego-Exo4D~\citep{grauman2024egoexo4d} defines object correspondence between synchronized ego and exo cameras, with co-segmentation and VOS baselines (XSegTx and XView-XMem~\citep{cheng2022xmem}) and PSALM~\citep{zhang2024psalm} as a zero-shot reference; later methods add language cues~\citep{fu2025objectrelator}, DINO-based mask matching~\citep{murlabadia2025omama}, long-term memory~\citep{hu2025lmeec}, cycle-consistent mask prediction~\citep{yan2026ccmp}, and multi-prompt experts on SAM2~\citep{pan2026v2sam}.
All train on Ego-Exo4D, whereas we evaluate 3AM zero-shot on ego-exo pairs as extreme viewpoint changes.

\begin{figure}[t]
    \centering
    \includegraphics[width=\linewidth]{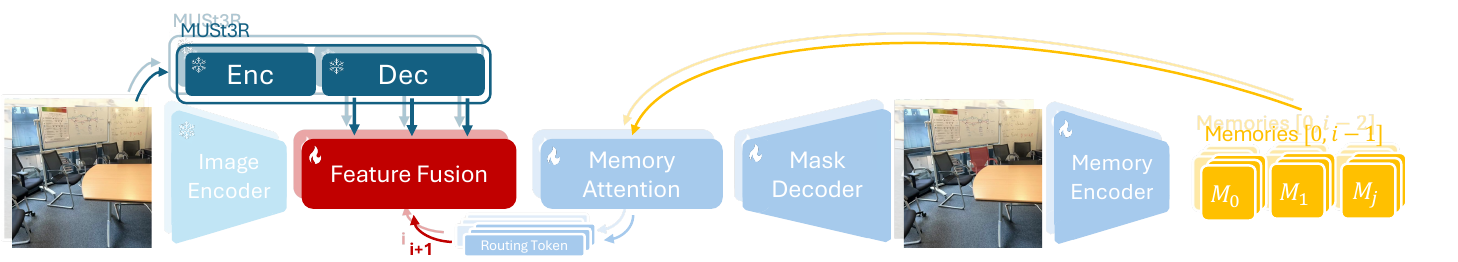}
    \caption{\textbf{3AM pipeline.}
    Frozen SAM2 and MUSt3R encode each frame, MUSt3R with its own memory of past frames.
    Feature Fusion (\cref{sec:fusion}) merges multi-layer MUSt3R features into the SAM2 feature used by memory attention, the mask decoder, and the memory encoder, so the memories $M_0,\dots,M_j$ are geometry-aware; routing tokens pass through memory attention, and their output $r^{i+1}$ sets the fusion of frame $i{+}1$.}
    \label{fig:pipeline}
\end{figure}

%% file: 03_method.tex
\section{Method}
\label{sec:method}
\label{sec:overview}
Given a video and a mask, point, or box prompt on one frame, the tracker outputs the object's mask in every frame where it is visible and an empty mask otherwise, as the object leaves the view, returns from another viewpoint, or moves.
In 3AM (\cref{fig:pipeline}), each frame $i$ is encoded by SAM2's frozen image encoder~\citep{ravi2024sam2}, giving an appearance feature $F_{\mathrm{2D}}$, and by frozen MUSt3R~\citep{cabon2025must3r}, an online reconstruction model with its own memory of past frames, giving geometric features $F_{\mathrm{3D}}$ that encode where the content lies.
Feature Fusion (\cref{sec:fusion}) merges them into $F_{\mathrm{fused}}$, which replaces $F_{\mathrm{2D}}$ in memory attention, the mask decoder, and the memory encoder, and routing tokens computed in memory attention at frame $i$ produce routing vectors $r^{i+1}$ that set the fusion of frame $i{+}1$.
Inference requires RGB only.
We keep SAM2 as the segmentation backbone instead of segmenting the predicted point cloud, which would inherit reconstruction errors, run offline, and forgo abundant 2D video data; geometry serves only as an association cue.

\subsection{What Do MUSt3R Features Encode?}
\label{sec:feature-analysis}

\begin{figure}[t]
    \centering
    \includegraphics[width=\linewidth]{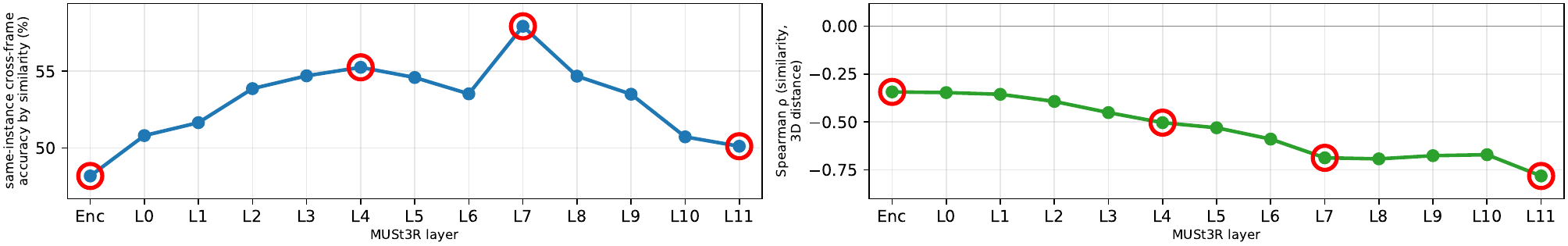}\\[1mm]
    \includegraphics[width=\linewidth]{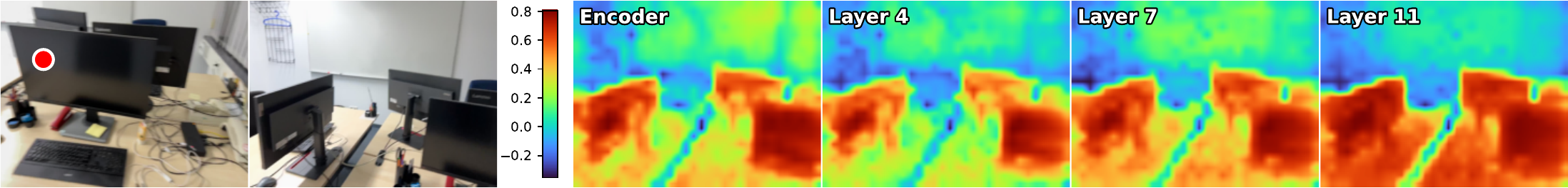}
    \caption{\textbf{MUSt3R features move from semantics to geometry with depth.}
    Top: per-layer statistics on ScanNet++ frame pairs (selected layers circled); same-instance accuracy peaks at middle layers, and the Spearman correlation with 3D distance grows with depth.
    Bottom: similarity of a red query on a monitor to another view; the encoder fires on every monitor, while geometry-bound deeper layers stop at the desk edge.}
    \label{fig:feature-analysis}
\end{figure}

MUSt3R predicts point maps in a common coordinate frame from an RGB stream, so its features encode cross-view correspondence, but which layers encode it in a form useful for object identity is unknown.
On pairs of ScanNet++ frames, we measure for the encoder output and the twelve decoder blocks (i) same-instance accuracy, the fraction of query patches whose most similar patch in the other frame belongs to the same instance, and (ii) the Spearman correlation between the cosine similarity and the 3D distance of two patches (\cref{fig:feature-analysis}).
The curves show one gradual transition.
Early layers are semantic, so a query on a monitor activates every monitor; intermediate decoder blocks attain the highest instance accuracy; deeper blocks correlate more with 3D distance and follow geometry rather than appearance, stopping at the desk edge.
As no single layer serves both roles, we take the encoder output and decoder blocks 4, 7, and 11 and let the model weight them.

\subsection{Feature Fusion with Routing Tokens}
\label{sec:fusion}

\begin{figure}[t]
    \centering
    \includegraphics[width=\linewidth]{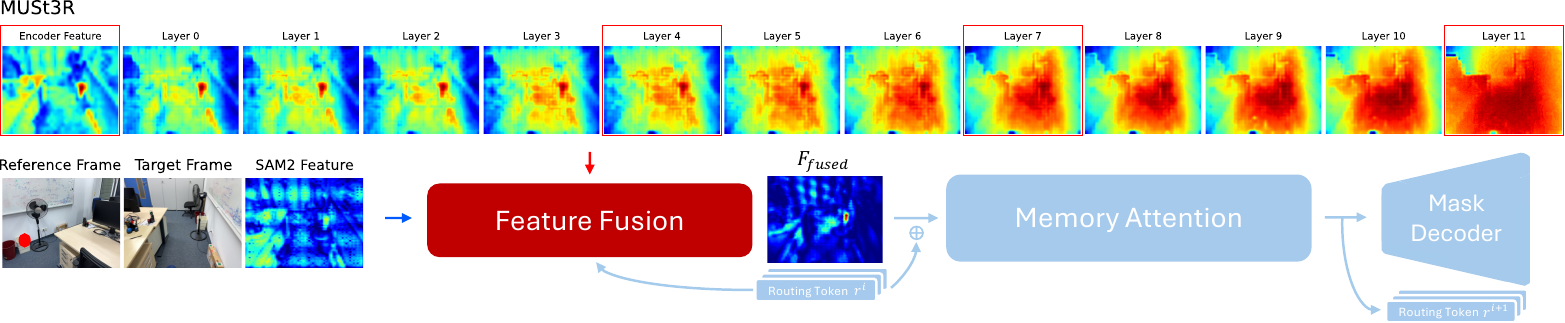}
    \caption{\textbf{Feature Fusion and routing.}
    Top: similarity of a reference-frame query point to the target frame for the MUSt3R encoder and decoder layers 0 to 11 (selected layers boxed).
    Bottom: the selected layers and the explicit geometry are projected, scaled and shifted per channel, gated, and added to the SAM2 feature with parameters decoded from the routing vector $r^{i}$; $F_{\mathrm{fused}}$ and the routing tokens are concatenated along the token dimension for memory attention, whose token output $r^{i+1}$ parameterizes the fusion of the next frame.}
    \label{fig:fusion}
\end{figure}

Let $F_{\mathrm{2D}}\in\mathbb{R}^{C\times H\times W}$ be the SAM2 feature of frame $i$ at stride 16 (frame index omitted).
The five geometric sources $F^{(k)}_{\mathrm{3D}}$, $k=1,\dots,5$, are the four selected MUSt3R layers and the frame's MUSt3R point map with its camera rays in Pl\"ucker coordinates, encoded by sinusoidal bands and a patch embedding.
A small convolutional head ($1{\times}1$ convolution, LayerNorm, GELU, $3{\times}3$ convolution) and bilinear upsampling project each source to $C\times H\times W$, giving $\hat{F}^{(k)}_{\mathrm{3D}}$, and the fusion is a gated, feature-wise modulated sum (\cref{fig:fusion}),
\begin{equation}
F_{\mathrm{fused}} = \gamma_0 \odot F_{\mathrm{2D}} + \beta_0 \;+\; \sum_{k=1}^{5} \alpha_k\,\big(\gamma_k \odot \hat{F}^{(k)}_{\mathrm{3D}} + \beta_k\big),
\qquad (\gamma_k,\beta_k,\alpha_k)=\mathrm{MLP}_k(r^{i}_k),
\label{eq:fusion}
\end{equation}
where $\gamma_k,\beta_k\in\mathbb{R}^{C}$ scale and shift each channel, $\alpha_k\in\mathbb{R}$ gates the whole source, $\odot$ is channel-wise multiplication, and the SAM2 feature ($k=0$) is not gated.
A two-layer $\mathrm{MLP}_k$ decodes $(\gamma_k,\beta_k,\alpha_k)$ from the routing vector $r^{i}_k\in\mathbb{R}^{256}$ that the routing tokens produced at the previous frame.
The routing vectors are the state that makes the fusion adaptive; each source enters under one global gate, which is cheap and can be switched off, and with all gates closed the model reduces to SAM2.

\paragraph{Routing Tokens.}
\label{sec:routing}
SAM2's memory attention has four layers, each with self-attention over the $HW$ frame tokens, cross-attention to the memory bank (spatial memories and object pointers of past frames), and a feed-forward layer.
We concatenate $N=12$ routing tokens, two per term $k=0,\dots,5$ of \cref{eq:fusion}, to the frame tokens along the token dimension and pass all tokens through these layers with shared weights; the routing tokens carry no position and receive no rotary encoding, and at every layer they see the fused frame, the memory bank, and the spatial tokens.
At frame $i$, each token is a learned embedding plus a linear projection of $r^{i}_k$; after the last layer, the two tokens of term $k$ are concatenated and projected to $r^{i+1}_k$, which sets $(\gamma_k,\beta_k,\alpha_k)$ of frame $i{+}1$.
A learned initial state $r^{0}_k$ serves the prompted frame, and $r^{1}_k=r^{0}_k$ plus the projected object pointer of the prompt conditions routing on the tracked object.
When the current 3D features agree with the object's stored memories, the geometric gates stay open; when the object has moved relative to the scene, memories and geometry disagree, and the gates close so that the model reverts to appearance matching, per frame and per source.
Since the memory encoder stores fused features, the memory bank keeps the geometric cue whenever it was used.
The routing output projection and the object-pointer projection are zero-initialized, so training starts as plain SAM2 with gates near zero and opens the geometric path where geometry reduces the loss.

\subsection{Training}
\label{sec:training}
We fine-tune SAM2.1-L on eight-frame clips with the SAM2 losses (focal and dice on the mask, $\ell_1$ on the predicted IoU, cross-entropy on object presence), training Feature Fusion, routing, memory attention, the memory encoder, and the mask decoder with the image encoder and MUSt3R frozen.

\paragraph{Data.}
Training mixes two static datasets with long camera trajectories, ScanNet++~\citep{yeshwanth2023scannet++} with masks projected from 3D labels and the synthetic ASE~\citep{avetisyan2024scenescript} with clean masks, and three dynamic video datasets, MOSE~\citep{MOSE}, MOSEv2~\citep{ding2025mosev2}, and SA-V~\citep{ravi2024sam2}.
Dynamic data are essential, as with static data alone the routing tokens never see geometry mislead and trust it unconditionally (\cref{sec:ablation}).
Static clips use \emph{anchor-then-contiguous} sampling, which pairs a prompt frame from anywhere in the sequence with a contiguous seven-frame block elsewhere, and dynamic clips use consecutive frames.

%% file: 04_experiments.tex
\section{Experiments}
\label{sec:exp}

\subsection{Setup}
\label{sec:setup}

\paragraph{Implementation.}
We build on SAM2.1-L ($1024^2$ input) and MUSt3R-512 (512 pixels on the long side); the fused feature has $C=256$ channels at $64\times64$, routing uses $N=12$ tokens and 256-dimensional vectors, and 3AM keeps SAM2's memory selection throughout.
Training uses AdamW with per-module learning rates between $10^{-6}$ and $10^{-5}$, cosine decay, and bfloat16 on one RTX~4090, with MUSt3R on a second GPU (details in \cref{app:details,app:eval,app:results}).

\paragraph{Benchmarks and Metrics.}
Conventional VOS and VOT benchmarks such as DAVIS~\citep{perazzi2017davis}, YouTube-VOS~\citep{xu2018youtube}, LVOS~\citep{hong2023lvos}, and LaSOT~\citep{fan2019lasot} mostly keep the object in view and the camera near its initial pose.
For large viewpoint changes we use ScanNet++~\citep{yeshwanth2023scannet++} and Replica~\citep{replica19arxiv}, whose trajectories cover whole rooms; 3D instance labels are projected into every frame with ground-truth poses, and the frame with the largest visible mask is the prompt, tracked forward and backward; our ScanNet++ training masks come from the same projection.
Besides the Whole Set of ScanNet++ objects, we filter a \emph{Selected Subset} of 286 objects whose tracks split into many disjoint visible segments, \ie frequent reappearance under large viewpoint changes.
\textbf{IoU} averages over all frames, including those where the object is absent (the tracking quality of VOTS, \citealp{kristan2023vots}), and thus penalizes false positives; \textbf{Tracking Recall}~\citep{lukezic2021performance} is the mean IoU over frames where the object is present; \textbf{Accuracy} is the mean IoU over present frames where the prediction overlaps the ground truth (the VOTS accuracy), \ie mask quality given correct localization.
Dynamic objects are evaluated on DAVIS17, LaSOT, LaSOT$_{\text{ext}}$, KITTI-MOTS~\citep{voigtlaender2019mots}, and EPIC-KITCHENS~\citep{damen2018epic,darkhalil2022visor}, objects that are moved and left in place on Aria Digital Twin (ADT)~\citep{pan2023aria}, cross-view correspondence on Ego-Exo4D~\citep{grauman2024egoexo4d} with the protocol of V$^2$-SAM~\citep{pan2026v2sam}, and 3D-projected tracks on class-agnostic ScanNet200.

\subsection{Tracking under Large Viewpoint Changes}
\label{sec:exp-wide}

\begin{table}[t]
\centering
\caption{\textbf{Tracking under large viewpoint changes and on dynamic benchmarks} (\%) for SAM2~\citep{ravi2024sam2}, SAM2Long~\citep{ding2024sam2long}, and DAM4SAM~\citep{videnovic2025distractor}; KITTI-MOTS and EPIC-KITCHENS use mask prompts; ADT: Aria Digital Twin~\citep{pan2023aria}. Best in red.}
\label{tab:main}
\scriptsize
\setlength{\tabcolsep}{1.6pt}
\resizebox{\textwidth}{!}{%
\begin{tabular}{@{}l ccc ccc ccc ccc ccc ccc ccc@{}}
\toprule
& \multicolumn{6}{c}{ScanNet++} & \multicolumn{3}{c}{\multirow{2}{*}{Replica}} & \multirow{2}{*}{DAVIS17} & \multirow{2}{*}{LaSOT} & \multirow{2}{*}{LaSOT$_{\text{ext}}$} & \multicolumn{3}{c}{\multirow{2}{*}{KITTI-MOTS}} & \multicolumn{3}{c}{\multirow{2}{*}{EPIC-KITCHENS}} & \multicolumn{3}{c}{\multirow{2}{*}{ADT}} \\
\cmidrule(lr){2-7}
& \multicolumn{3}{c}{Whole Set} & \multicolumn{3}{c}{Selected Subset} & \multicolumn{3}{c}{} & & & & \multicolumn{3}{c}{} & \multicolumn{3}{c}{} & \multicolumn{3}{c}{} \\
\cmidrule(lr){2-4}\cmidrule(lr){5-7}\cmidrule(lr){8-10}\cmidrule(lr){11-11}\cmidrule(lr){12-12}\cmidrule(lr){13-13}\cmidrule(lr){14-16}\cmidrule(lr){17-19}\cmidrule(lr){20-22}
Method & IoU & Recall & Acc. & IoU & Recall & Acc. & IoU & Recall & Acc. & $\mathcal{J}\&\mathcal{F}$ & AUC & AUC & IoU & Recall & Acc. & IoU & Recall & Acc. & IoU & Recall & Acc. \\
\midrule
SAM2       & 81.2 & 32.5 & 69.3 & 74.3 & 28.2 & 66.6 & 80.0 & 56.0 & 69.0 & 89.0 & 70.0 & 56.9 & 81.9 & 83.8 & 84.0 & 78.5 & \best{83.1} & \best{86.8} & 81.3 & 68.1 & 71.9 \\
SAM2Long   & 81.3 & 41.7 & 69.6 & 78.3 & 36.1 & 62.4 & 76.9 & 52.0 & 62.7 & 88.3 & 73.9 & -- & -- & -- & -- & -- & -- & -- & 81.2 & 68.3 & 72.9 \\
DAM4SAM    & 83.1 & 41.9 & 67.8 & 76.6 & 37.5 & 63.8 & 79.3 & 58.5 & 67.8 & 86.6 & 75.1 & \best{60.9} & 81.9 & 83.8 & \best{84.7} & 77.7 & 83.0 & 86.4 & 81.0 & 69.1 & 73.1 \\
\midrule
3AM (Ours) & \best{89.5} & \best{59.3} & \best{71.6} & \best{90.6} & \best{65.7} & \best{72.4} & \best{83.4} & \best{67.8} & \best{70.9} & \best{90.4} & \best{75.4} & 60.1 & \best{82.6} & \best{83.9} & 84.0 & \best{78.8} & \best{83.1} & 86.7 & \best{85.3} & \best{76.9} & \best{75.5} \\
\bottomrule
\end{tabular}}
\end{table}

\begin{figure}[t]
    \centering
    \includegraphics[width=\textwidth]{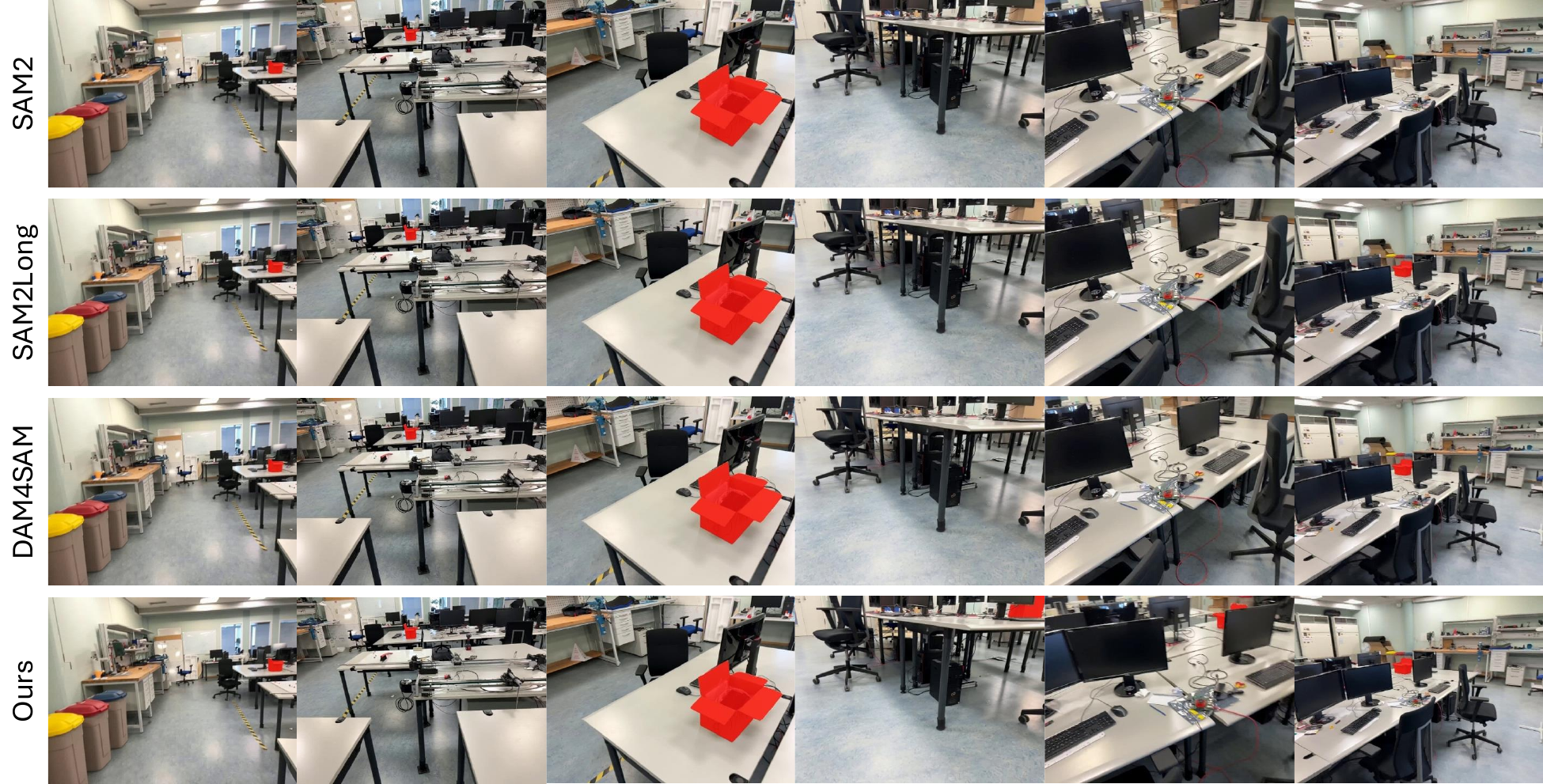}
    \caption{\textbf{Qualitative comparison.} The leftmost frame is prompted. SAM2-based trackers drift to distractors or return nothing after the object leaves the view; 3AM recovers the same instance from new viewpoints.}
    \label{fig:vis_comp}
\end{figure}

\begin{table}[t]
\centering
\begin{minipage}[t]{0.605\textwidth}
\centering
\caption{\textbf{Comparison with other tracker families} on the ScanNet++ Selected Subset (\%). Prompt.: promptable; Online: streaming; RGB: no poses, depth, or per-video optimization; $\dagger$: SAM2 fine-tuned on our data. Best in red.}
\label{tab:baselines-3d}
\scriptsize
\setlength{\tabcolsep}{5pt}
\renewcommand{\arraystretch}{0.95}
\resizebox{\linewidth}{!}{%
\begin{tabular}{@{}l ccc ccc@{}}
\toprule
Method & Prompt. & Online & RGB & IoU & Recall & Acc. \\
\midrule
\multicolumn{7}{@{}l}{\emph{Memory-based VOS}} \\
SAM2~\citep{ravi2024sam2}                    & \cmark & \cmark & \cmark & 74.3 & 28.2 & 66.6 \\
SAM2$^\dagger$                               & \cmark & \cmark & \cmark & 77.5 & 40.9 & 67.0 \\
DAM4SAM~\citep{videnovic2025distractor}      & \cmark & \cmark & \cmark & 76.6 & 37.5 & 63.8 \\
SAM~3~\citep{carion2025sam3}                 & \cmark & \cmark & \cmark & 76.8 & 31.6 & 60.7 \\
\midrule
\multicolumn{7}{@{}l}{\emph{Point and 4D tracking, masks decoded by SAM2}} \\
CoTracker3~\citep{karaev2024cotracker3}+SAM2 & \cmark & \cmark & \cmark & 73.6 & 26.7 & 66.7 \\
OmniMotion~\citep{wang2023omnimotion}+SAM2   & \cmark & \xmark & \xmark & 58.0 & 9.3  & 56.4 \\
4RC~\citep{luo2026fourrc}+SAM2               & \cmark & \xmark & \cmark & 80.6 & 41.3 & 51.7 \\
\midrule
\multicolumn{7}{@{}l}{\emph{3D-aware segmentation and geometric matching}} \\
PanSt3R~\citep{zust2025panst3r}              & \xmark & \xmark & \cmark & 52.1 & 7.1  & 9.2 \\
SegMASt3R~\citep{Jayanti2025SegMASt3R}       & \xmark & \xmark & \cmark & 54.1 & 35.1 & 40.4 \\
EmbodiedSAM~\citep{xu2025embodiedsam}                      & \xmark & \cmark & \xmark & 70.7 & 16.9 & 28.1 \\
MUSt3R-Thr+SAM2 (training-free)              & \cmark & \cmark & \cmark & 65.1 & 30.0 & 31.0 \\
MUSt3R-Reproj+SAM2 (training-free)           & \cmark & \cmark & \cmark & 81.5 & 40.7 & 50.4 \\
IGGT4D~\citep{zou2026iggt4d}                 & \xmark & \cmark & \cmark & 78.1 & 52.6 & 57.9 \\
\midrule
3AM (Ours) & \cmark & \cmark & \cmark & \best{90.6} & \best{65.7} & \best{72.4} \\
\bottomrule
\end{tabular}}
\end{minipage}\hfill
\begin{minipage}[t]{0.365\textwidth}
\centering
\caption{\textbf{Routing shares} (\%) of the appearance terms (SAM2 feature, MUSt3R encoder) and geometric terms (decoder blocks 4, 7, 11, explicit geometry) on static ScanNet++ and dynamic DAVIS17.}
\label{tab:routing-shares}
\scriptsize
\begin{tabular*}{\linewidth}{@{\extracolsep{\fill}}l cc@{}}
\toprule
Terms & Static & Dynamic \\
\midrule
Appearance & 49.3 & 82.2 \\
Geometric  & 50.7 & 17.8 \\
\bottomrule
\end{tabular*}\\[5pt]
\caption{\textbf{Efficiency} on an RTX~4090 ($1024^2$ input, MUSt3R at 512), single-object tracking.}
\label{tab:efficiency}
\scriptsize
\setlength{\tabcolsep}{3pt}
\resizebox{\linewidth}{!}{%
\begin{tabular}{@{}l cc@{}}
\toprule
Method & FPS & Mem.\ (GB) \\
\midrule
SAM2~\citep{ravi2024sam2}               & 14.4 & 3.6 \\
SAM2Long~\citep{ding2024sam2long}       & 7.1  & 6.8 \\
DAM4SAM~\citep{videnovic2025distractor} & 12.3 & 3.5 \\
3AM (Ours)                              & 10.2 & 4.3 \\
\bottomrule
\end{tabular}}
\end{minipage}
\end{table}

\cref{tab:main} compares 3AM with SAM2~\citep{ravi2024sam2} and its memory-selection extensions SAM2Long~\citep{ding2024sam2long} and DAM4SAM~\citep{videnovic2025distractor}, \cref{tab:baselines-3d} adds SAM2 fine-tuned on our data (SAM2$^\dagger$), which separates the data from the architecture, and other tracker families, and \cref{fig:vis_comp} shows examples.
SAM2 recovers few objects after they leave the view (28.2 Recall on the Selected Subset), SAM~3~\citep{carion2025sam3} and memory selection stay below 38, and SAM2$^\dagger$ reaches only 40.9, since memory attention without a geometric cue cannot separate the target from look-alikes under a new viewpoint.
Point-based baselines track points sampled in the prompt mask with CoTracker3~\citep{karaev2024cotracker3}, OmniMotion~\citep{wang2023omnimotion}, or the 4D model 4RC~\citep{luo2026fourrc} and prompt SAM2 with them; point trackers cannot re-identify the object after a long absence, OmniMotion needs per-video optimization, and 4RC reaches 41.3 Recall but runs offline and yields coarse masks (51.7 Accuracy).
The 3D-aware baselines are consistent by construction but are not online promptable trackers, since PanSt3R~\citep{zust2025panst3r} predicts panoptic segments offline, SegMASt3R~\citep{Jayanti2025SegMASt3R} matches segments between two views, EmbodiedSAM~\citep{xu2025embodiedsam} segments 3D instances online from posed RGB-D, and the training-free MUSt3R-Thr+SAM2, which prompts SAM2 with the thresholded MUSt3R feature similarity to the prompt mask, is not identity-aware.
Reprojecting the prompt mask through MUSt3R's point maps (MUSt3R-Reproj+SAM2) reaches 40.7 Recall but yields coarse masks (50.4 Accuracy).
IGGT4D supports online inference but relies on instance-feature clustering without native prompt-conditioned segmentation.
We cannot compare with G$^2$TAM~\citep{zhu2026g2tam}, whose code has not been released.
3AM is best on all datasets and metrics, with 90.6 IoU and 65.7 Recall on the Selected Subset, +12.3 IoU over SAM2Long and +28.2 Recall over DAM4SAM.
3AM also improves on the Whole Set, and the advantage holds for point and box prompts and for unordered photos, which are first reordered by feature retrieval.

\subsection{Dynamic Objects and Routing}
\label{sec:exp-dynamic}
The geometric prior must not hurt moving objects.
On benchmarks dominated by object motion and stable cameras, including egocentric hand-object manipulation in EPIC-KITCHENS (\cref{tab:main}, right), 3AM matches or exceeds SAM2 everywhere (90.4 versus 89.0 on DAVIS17, 75.4 versus 70.0 on LaSOT) and is on par with the distractor-aware DAM4SAM, as the routing tokens close the geometric gates when the current geometry disagrees with the memories.
The lower part of \cref{fig:in-the-wild} shows the case that neither regime alone solves, a bottle carried across a room, put down, and revisited after the camera has left; both trackers follow it in the hand, but only 3AM re-identifies it on every return, even when small and distant.
On ADT, where many objects are moved and then left in place, 3AM reaches 76.9 Recall, 7.8 points above the best SAM2-based tracker (\cref{tab:main}).

\begin{table}[t]
\centering
\begin{minipage}[t]{0.575\textwidth}
\centering
\caption{\textbf{Ego-exo object correspondence on Ego-Exo4D} (IoU, \%). Tr.: trained on Ego-Exo4D; SAM2 and 3AM are zero-shot. Baselines from \citet{pan2026v2sam}; best in red.}
\label{tab:egoexo}
\scriptsize
\setlength{\tabcolsep}{3pt}
\resizebox{\linewidth}{!}{%
\begin{tabular}{@{}l c ccc@{}}
\toprule
Method & Tr. & Ego$\rightarrow$Exo & Exo$\rightarrow$Ego & Mean \\
\midrule
PSALM~\citep{zhang2024psalm} (zero-shot)      & \xmark & 7.4  & 2.1  & 4.8 \\
CMX~\citep{zhang2023cmx}                       & \cmark & 6.8  & 12.0 & 9.4 \\
XSegTx~\citep{grauman2024egoexo4d}             & \cmark & 18.9 & 27.1 & 23.0 \\
XView-XMem~\citep{grauman2024egoexo4d}         & \cmark & 19.3 & 16.6 & 18.0 \\
XMem + XSegTx~\citep{grauman2024egoexo4d}      & \cmark & 34.9 & 25.0 & 30.0 \\
ObjectRelator~\citep{fu2025objectrelator}      & \cmark & 35.3 & 40.3 & 37.8 \\
O-MaMa~\citep{murlabadia2025omama}             & \cmark & 42.6 & 44.1 & 43.4 \\
V$^2$-SAM~\citep{pan2026v2sam} (single expert) & \cmark & 44.5 & 47.3 & 45.9 \\
V$^2$-SAM~\citep{pan2026v2sam} (multi-expert)  & \cmark & 46.3 & 49.6 & 48.0 \\
\midrule
SAM2 (zero-shot)                               & \xmark & 14.0 & 20.6 & 17.3 \\
3AM (Ours, zero-shot)                          & \xmark & \best{49.1} & \best{56.0} & \best{52.6} \\
\bottomrule
\end{tabular}}
\end{minipage}\hfill
\begin{minipage}[t]{0.395\textwidth}
\centering
\caption{\textbf{Projected 3AM tracks as class-agnostic 3D instances on ScanNet200}, a consistency check rather than a 3D segmentation method; all methods use ground-truth depth and poses. Best in red.}
\label{tab:scannet200}
\scriptsize
\setlength{\tabcolsep}{3pt}
\resizebox{\linewidth}{!}{%
\begin{tabular}{@{}l ccc@{}}
\toprule
Method & AP & AP$_{50}$ & AP$_{25}$ \\
\midrule
\multicolumn{4}{@{}l}{\emph{Offline}} \\
SAMPro3D~\citep{xu2025sampro3d}       & 18.0 & 32.8 & 56.1 \\
Open3DIS~\citep{nguyen2024open3dis}   & 34.6 & 43.1 & 48.5 \\
SAI3D~\citep{yin2024sai3d}            & 28.2 & 47.2 & 67.9 \\
SAM2Object~\citep{zhao2025sam2object} & 34.0 & 52.7 & 70.3 \\
\midrule
\multicolumn{4}{@{}l}{\emph{Online}} \\
SAM3D~\citep{yang2023sam3d}           & 20.2 & 35.7 & 55.5 \\
EmbodiedSAM~\citep{xu2025embodiedsam}               & 42.2 & 63.7 & \best{79.6} \\
3AM (Ours)                            & \best{47.3} & 59.7 & 75.3 \\
\bottomrule
\end{tabular}}
\end{minipage}
\end{table}

\paragraph{What Do the Routing Tokens Learn?}
\label{sec:exp-routing}
On static ScanNet++, geometric terms receive half of the routing-coefficient magnitude (50.7\%, \cref{tab:routing-shares}), whereas on dynamic DAVIS17 they receive 17.8\%; decoder blocks 4 and 11 carry most of the static share, their signed gains flip or vanish on dynamic data, and the first 100 frames after the prompt give the same shares.
Training the fusion without routing, with $(\gamma_k,\beta_k,\alpha_k)$ learned as constants shared by all frames, not set by hand (\cref{tab:ablation}c), keeps most of the wide-baseline gain (59.6 versus 65.7 Recall) but loses 3.5 points on DAVIS17.

\subsection{Cross-View Correspondence on Ego-Exo4D}
\label{sec:exp-egoexo}
An egocentric and an exocentric view of one object form an extreme viewpoint change.
Following V$^2$-SAM~\citep{pan2026v2sam}, the target-view mask is predicted from the query-view image and mask in both directions; 3AM treats the pair as a two-frame video prompted on the query view, with MUSt3R initialized on both views.
Without Ego-Exo4D training data, unlike all other methods in \cref{tab:egoexo}, 3AM reaches 52.6 mean IoU, 4.6 points above the best trained method and 35.3 above zero-shot SAM2, with the larger margin for Exo$\rightarrow$Ego (56.0 versus 49.6).

\subsection{Ablation Study}
\label{sec:ablation}

\begin{table}[t]
\centering
\begin{minipage}[c]{0.5445\textwidth}
\centering
\scriptsize
\setlength{\tabcolsep}{3pt}
\resizebox{\linewidth}{!}{%
\begin{tabular}{@{}l l cccc@{}}
\toprule
& Variant & IoU & Recall & Acc. & DAVIS17 \\
\midrule
& 3AM (full)                                  & 90.6 & 65.7 & 72.4 & 90.4 \\
\midrule
(a) & SAM2 features only (SAM2$^\dagger$)      & 77.5 & 40.9 & 67.0 & 87.3 \\
(b) & MUSt3R features only                     & 84.6 & 52.7 & 61.3 & 84.9 \\
\midrule
(c) & No routing, trained fixed fusion         & 89.1 & 59.6 & 68.5 & 86.9 \\
(d) & Cross-attention merger, +103M parameters & 91.2 & 66.8 & 72.5 & 90.5 \\
\midrule
(e) & Static data only (ScanNet++, ASE)        & 92.3 & 68.5 & 73.6 & 80.2 \\
(f) & Dynamic data only (MOSE, MOSEv2, SA-V)   & 76.6 & 45.7 & 64.3 & 92.8 \\
\midrule
(g) & Consecutive-frame sampling               & 89.6 & 63.8 & 70.1 & 90.5 \\
\bottomrule
\end{tabular}}
\end{minipage}\hfill
\begin{minipage}[c]{0.4255\textwidth}
\includegraphics[width=\linewidth]{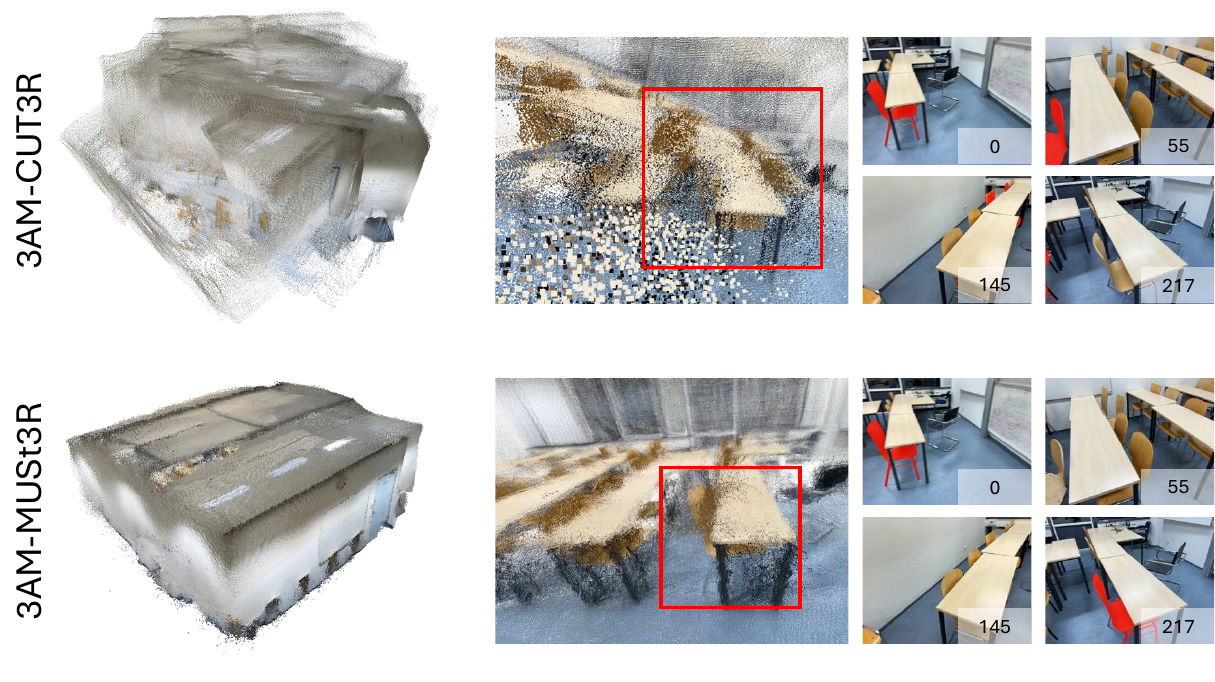}
\end{minipage}\\[2pt]
\begin{minipage}[t]{0.5445\textwidth}
\captionsetup{position=bottom}
\captionof{table}{\textbf{Ablations} on the ScanNet++ Selected Subset (\%) and DAVIS17 ($\mathcal{J}\&\mathcal{F}$): feature streams (a, b), routing (c), fusion design (d), training data (e, f), and clip sampling for static scenes (g).}
\label{tab:ablation}
\end{minipage}\hfill
\begin{minipage}[t]{0.4255\textwidth}
\captionsetup{position=bottom}
\captionof{figure}{\textbf{CUT3R vs.\ MUSt3R geometry.} CUT3R (top) places the same table at inconsistent 3D locations; MUSt3R (bottom) keeps objects aligned, which its features inherit.}
\label{fig:comp_3r}
\end{minipage}
\end{table}

\paragraph{Feature Streams and Fusion Design.}
Both streams are required (\cref{tab:ablation}a, b).
SAM2 features alone reach 40.9 Recall, lacking a geometric anchor under new viewpoints, and MUSt3R features alone localize the object (52.7 Recall) but give coarser masks (61.3 versus 72.4 Accuracy) without SAM2's appearance detail.
A hierarchical cross-attention merger (d) gains at most 1.1 points while adding 103M parameters and attention over all positions per source; the lightweight fusion suffices because the two feature maps are spatially aligned, and its scalar gates are what routing switches.

\paragraph{Training Data and Sampling.}
Static data only (e) trusts geometry everywhere, the best variant on ScanNet++ (68.5 Recall) but 10.2 points worse on DAVIS17, and dynamic data only (f) never opens the geometric path, the best on DAVIS17 (92.8) but at 45.7 Recall on ScanNet++, so routing needs both domains to learn when geometry is reliable.
Consecutive frames (g) give fewer reappearances; anchor-then-contiguous sampling adds 1.9 Recall and 2.3 Accuracy at no cost on DAVIS17.

\paragraph{3D Backbone.}
As 3D feature sources, VGGT and $\pi^3$ attend over all frames jointly and cannot run online on long sequences, and the online CUT3R drifts (\cref{fig:comp_3r}) and gives features too coarse for re-identification, reaching 27.5 Recall against 65.7 with MUSt3R.
MUSt3R keeps objects aligned, and its frozen features are the strongest correspondence source among online models (\cref{app:feature-analysis}).

\paragraph{Efficiency.}
\label{sec:efficiency}
3AM fine-tunes 22.455M parameters and runs at 10.2 FPS with 4.3~GB of peak memory, faster and lighter than SAM2Long (\cref{tab:efficiency}); MUSt3R dominates the extra time, and the cost stays bounded on long videos as SAM2's memory bank is fixed and MUSt3R's memory, which otherwise grows with the sequence, discards old entries at inference beyond a fixed budget.

\subsection{Lifting Consistent Tracks to 3D}
\label{sec:exp-3d}
As a final check that the identity preserved in 2D is consistent in 3D, we propagate SAM2 automatic-mask proposals from keyframes with 3AM, associate tracks by temporal overlap, and project them into 3D with ground-truth depth and poses instead of MUSt3R's geometry, since the annotations are defined on the ground-truth 3D scans.
Lifting methods must merge per-view masks in 3D because the masks are inconsistent, whether from 2D proposals~\citep{jung2025details,zhao2025sam2object} or 3D detectors~\citep{boudjoghra2025openyolo,nguyen2024open3dis}, whereas consistent tracks need little merging.
Under the class-agnostic ScanNet200 protocol (\cref{tab:scannet200}), the projected tracks reach 47.3 AP, on par with or above offline~\citep{xu2025sampro3d,nguyen2024open3dis,yin2024sai3d,zhao2025sam2object} and online~\citep{yang2023sam3d,xu2025embodiedsam} methods that merge in 3D.

%% file: 05_conclusion.tex
\section{Conclusion}
\label{sec:conclusion}
We presented 3AM, a promptable tracker that preserves object identity through reappearance and large viewpoint changes while running online on RGB alone, without poses, depth, or offline reconstruction.
Coupling SAM2 with the online reconstruction model MUSt3R supplies the location cue that appearance matching lacks, and because this cue holds only while the object stays in place, routing tokens in SAM2's memory attention decide per frame how much the fusion relies on geometry.
One model thus re-identifies static objects from new viewpoints, follows objects that move, transfers zero-shot to ego-exo correspondence, and produces tracks that stay consistent in 3D.

%% file: 10_appendix.tex
\section{Implementation Details}
\label{app:details}

\subsection{SAM2 Backbone}
All experiments use SAM2.1-L.
A Hiera~\citep{ryali2023hiera} image encoder produces multi-scale features; the stride-16 and stride-32 outputs are fused by an FPN~\citep{lin2017featureFPN} into the frame embedding ($256\times64\times64$ for a $1024^2$ input), while the stride-4 and stride-8 features are injected only into the mask decoder to recover fine boundaries.
Memory attention has four layers with sinusoidal absolute positional embeddings and 2D RoPE on the spatial tokens; object-pointer tokens do not use RoPE.
The prompt encoder follows SAM, and the mask decoder produces the object pointer stored in the memory bank, an object-presence score, and several mask candidates, from which the one with the highest predicted IoU is selected when the prompt is ambiguous.
The memory encoder projects features to 64 channels; the 256-dimensional object pointer is split into four 64-dimensional tokens for cross-attention.
The memory bank holds the prompted frame and up to six recent frames, plus up to 16 object pointers.
For multi-object videos, the image encoder is shared and each object keeps its own memory bank.

\subsection{Feature Fusion and Routing}
\label{app:fusion-details}
\paragraph{MUSt3R Features.}
We use MUSt3R-512 with its own view-coverage based memory selection; since this memory otherwise grows with the sequence, at inference we discard old entries once it exceeds a fixed budget.
From each frame we take the encoder output (1024 channels) and the outputs of decoder blocks 4, 7, and 11 (768 channels each), all at stride 16.
Block 11 is the last decoder block, whose normalized output feeds the point-map head; we also read the head output (point map, confidence, and the camera rays derived from the estimated focal length) for the explicit geometry source.

\paragraph{Projection Heads.}
Each of the four MUSt3R features passes through a $1{\times}1$ convolution to 384 channels, a channel-wise LayerNorm, GELU, and a $3{\times}3$ convolution to 256 channels, followed by bilinear upsampling to $64\times64$.
The explicit geometry source concatenates the point map, encoded with a sinusoidal band encoding of six frequencies (39 channels), and the Pl\"ucker coordinates of the camera rays (6 channels), and applies a patch embedding (kernel and stride 16) to 256 channels, followed by the same upsampling.

\paragraph{Modulation MLPs.}
For each geometric source $k=1,\dots,5$, a two-layer $\mathrm{MLP}_k$ ($256\rightarrow512\rightarrow513$, GELU) maps $r^{i}_k$ to the per-channel scale and shift $\gamma_k,\beta_k\in\mathbb{R}^{256}$ and the scalar gate $\alpha_k$.
$\mathrm{MLP}_0$ ($256\rightarrow512\rightarrow512$) produces $\gamma_0,\beta_0$ for the SAM2 feature; its output layer is zero-initialized so that this term starts as the identity.

\paragraph{Routing Tokens.}
There are six terms $k=0,\dots,5$ (the SAM2 feature, four MUSt3R layers, and the explicit geometry source) with two routing tokens each, $N=12$ in total.
Each token is a learned embedding plus a linear projection ($256\rightarrow256$) of $r^{i}_k$.
Routing tokens are concatenated to the $HW=4096$ spatial tokens and pass through the four memory-attention layers with the shared self-attention, cross-attention, and feed-forward weights; in cross-attention the rotary encoding is applied to the spatial queries only.
After the last layer, the two tokens of term $k$ are concatenated and projected ($512\rightarrow256$) to $r^{i+1}_k$, used by frame $i{+}1$.
This output projection, the initial states $r^{0}_k$, and the projection of the prompt object pointer ($256\rightarrow6\times256$) are zero-initialized, so training starts from $r^{i}_k=r^{0}_k$ with gates near zero.
The prompted frame is fused with $r^{0}_k$ and its fused feature is stored in memory; every later frame is fused with the routing vectors produced by the previous frame, and the memory encoder stores the fused feature of that frame together with its predicted mask.

\subsection{Training Details}
\label{app:training-details}
\paragraph{Losses and Optimizer.}
We follow the SAM2 loss design with focal loss (weight 20) and dice loss (weight 1) on the mask, an $\ell_1$ loss on the predicted IoU (weight 1), and a cross-entropy loss on object presence (weight 1).
Trainable parameters (22.455M in total) are the Feature Fusion module, the routing tokens and projections, memory attention, the memory encoder, and the mask decoder; the image encoder, its high-resolution projection convolutions, and MUSt3R are frozen.
We use AdamW with learning rates of $5{\times}10^{-6}$ (fusion and decoder), $10^{-5}$ (memory attention), and $10^{-6}$ (memory encoder), cosine decay to one tenth of the initial rate, gradient clipping at norm 1, bfloat16 autocast, and one clip of eight frames per iteration.
Training runs on one RTX~4090; MUSt3R inference for the clip runs in worker processes on a second GPU.

\paragraph{Prompts.}
On ScanNet++ the prompt is always a mask, because the projected 2D masks are too noisy for reliable point sampling and the model should learn from the mask directly.
On ASE the prompt is a point without iterative corrections.
On MOSE, MOSEv2, and SA-V we keep SAM2's default protocol, with point, box, or mask prompts on the first frame and correction clicks on up to two frames.

\paragraph{Clip Sampling for Static Scenes.}
A training clip must contain a reappearance under a large viewpoint change, but SAM2's memory holds few frames, and frames sampled at random from a large object may show disjoint parts of it, contradicting the goal of locating the object by its 3D position.
For ASE and ScanNet++ we therefore use anchor-then-contiguous sampling, in which the prompt frame is drawn from anywhere in the sequence and the remaining seven frames form a contiguous block elsewhere, before or after the anchor.
The anchor supplies the viewpoint jump, and the block retains temporally coherent motion and matches the inference regime.
Fully random frames would expose the model to more viewpoint variation, but the resulting model attends at inference to non-conditioning memories that are wrong, which degrades the results.
The sequence is the list of keyframes for which masks and cached MUSt3R features exist (ScanNet++ frames at stride 10; ASE keyframes selected by co-visibility).
The prompted frame is drawn uniformly among frames in which the target is visible with sufficient area, and the remaining seven frames form a block of consecutive keyframes drawn uniformly among valid block starts elsewhere in the sequence, before or after the anchor.
The block is kept in forward temporal order and is not filtered for target visibility, so the target may be absent in part or all of the block, which supervises absence prediction.
For MOSE, MOSEv2, and SA-V the eight frames are consecutive.

\paragraph{MUSt3R Features During Training.}
For the static datasets we precompute MUSt3R features by streaming the full keyframe sequence of every scene once and caching the selected layers, so that training sees the same long-term features as inference; contiguous frames also keep these features consistent, whereas frames far apart can be reconstructed poorly.
Because at inference MUSt3R only sees the frames it is given, 20\% of the ScanNet++ clips and 10\% of the ASE clips are instead processed online over the eight sampled frames.
Dynamic datasets are always processed online.

\subsection{Datasets}
\label{app:datasets}

\begin{table}[h]
\centering
\caption{\textbf{Training data.}}
\label{tab:datasets}
\small
\resizebox{\textwidth}{!}{
\begin{tabular}{lccccc}
\toprule
& ScanNet++ & ASE & MOSE & MOSEv2 & SA-V \\
\midrule
Real           & \cmark & \xmark & \cmark & \cmark & \cmark \\
Dynamic        & \xmark & \xmark & \cmark & \cmark & \cmark \\
Clip sampling  & Anchor + contiguous & Anchor + contiguous & Consecutive & Consecutive & Consecutive \\
Prompt         & Mask & Point & Point / box / mask & Point / box / mask & Point / box / mask \\
MUSt3R features & Cached (20\% online) & Cached (10\% online) & Online & Online & Online \\
\bottomrule
\end{tabular}
}
\end{table}

\paragraph{ScanNet++~\citep{yeshwanth2023scannet++}} is a high-quality indoor RGB-D dataset with accurate poses, dense trajectories, and 3D instance annotations.
Its 2D masks are projections of the 3D labels, so they carry reprojection noise from depth and pose errors, and the pose annotation omits stretches of frames that the SfM pipeline discards.
It nevertheless provides the realistic geometry and viewpoint diversity needed to learn view-consistent identification.

\paragraph{Aria Synthetic Environments (ASE)~\citep{avetisyan2024scenescript}} contains procedurally generated indoor scenes rendered with simulated Aria sensors, with clean 2D and 3D annotations.
We use scenes whose number of views lies in a practical range, since too few views give little viewpoint variation and too many exceed the memory of MUSt3R.

\paragraph{MOSE~\citep{MOSE}, MOSEv2~\citep{ding2025mosev2}, and SA-V~\citep{ravi2024sam2}} are video segmentation datasets with heavy occlusion, disappearance and reappearance, small objects, and, in SA-V, a very large number of videos.
They keep the appearance-based VOS ability of the model and, more importantly, supply moving objects, which the routing tokens need in order to learn when geometry is unreliable.
They also prevent the model from hallucinating tracks from geometry alone.

\section{Evaluation Details}
\label{app:eval}

\subsection{ScanNet++ and Replica}
\label{app:eval-scannetpp}
For each scene we project the 3D instance labels into every frame with the ground-truth poses and depth, select the frame with the largest visible mask of an instance as the prompt frame, and track forward and backward.
Objects that are absent in the first frame are therefore still evaluated from a valid prompt.
We construct the Selected Subset of ScanNet++ ourselves: a visible segment is a maximal run of consecutive frames in which the instance is visible, and an instance is selected when more than eight of its visible segments last at least 12 frames.
\cref{tab:subset-stats} summarizes both groups.
Selected instances disappear more often, stay absent longer, and span most of the sequence, which stresses re-identification after long absences.

\begin{table}[h]
\centering
\caption{\textbf{ScanNet++ Selected Subset statistics}, median (25th--75th percentile) per instance. The fraction of empty frames, in which the object is absent, is 0.6961 on the Selected Subset and 0.7871 on the Whole Set.}
\label{tab:subset-stats}
\small
\resizebox{\textwidth}{!}{
\begin{tabular}{lcccc}
\toprule
Set & \makecell{\# disjoint visible segments\\with $>$12 consecutive frames} & \makecell{\# disjoint visible\\segments} & \makecell{\# absent frames} & \makecell{Frame coverage from\\first to last appearance} \\
\midrule
Selected ($N{=}286$)     & 9 (8--10) & 20 (14--26) & 552 (384--696) & 0.948 (0.908--0.981) \\
Non-selected ($N{=}3800$) & 3 (2--5)  & 8 (5--13)   & 367 (227--545) & 0.787 (0.565--0.933) \\
\bottomrule
\end{tabular}
}
\end{table}

\paragraph{Overlap with MUSt3R Training Data.}
MUSt3R was trained on, among other datasets, a ScanNet++ split.
MUSt3R is used as a frozen feature extractor only, none of our validation scenes appear in MUSt3R's training split or in our own training data, and our evaluation asks for online promptable tracking over thousand-frame videos rather than the short multi-view tuples MUSt3R was trained on.

\paragraph{Multiple Objects.}
As in SAM2, the image encoder, MUSt3R, and the Feature Fusion inputs are shared across objects in the same video; each object keeps its own memory bank and its own routing state, so the per-object cost of 3AM is the cost of memory attention and the mask decoder only.

\subsection{Viewpoint Analysis}
\label{app:eval-viewpoint}
\begin{figure}[h]
    \centering
    \includegraphics[width=\textwidth]{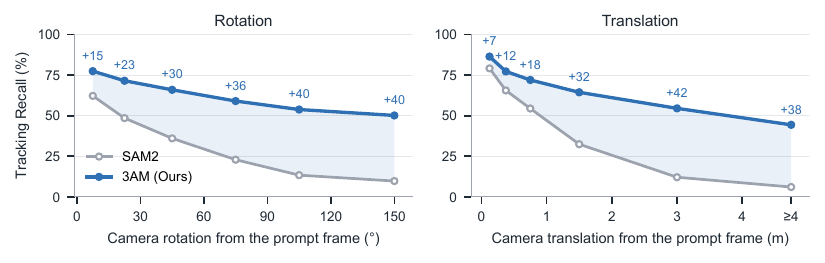}
    \vspace{-3mm}
    \caption{\textbf{Tracking Recall versus camera-pose change from the prompt frame} on the 286 objects of the ScanNet++ Selected Subset.
    Frames are binned by the rotation (left) or translation (right) of the camera relative to the prompt frame; each bin averages frames within an object and then objects.
    Numbers above the curve give the gain of 3AM over SAM2 in points; all gains have scene-bootstrap 95\% confidence intervals excluding zero (\cref{app:eval-viewpoint}).}
    \label{fig:analysis}
\end{figure}

For the breakdown in \cref{fig:analysis}, both trackers are initialized with the same maximum-area ground-truth mask and track all RGB frames in both directions; the prompt frame is excluded from the scores.
Camera translation is the metric distance between the COLMAP camera centres of the current and the prompt frame, and rotation is the geodesic angle between their orientations; both describe camera-pose changes rather than object-centric viewing angles.
Recall in a bin is the mean IoU over GT-positive frames in that bin, averaged within an object and then over objects; the number of objects with frames in a bin ranges from 285 (0--15$^\circ$) to 199 (120--180$^\circ$) and from 285 (0--0.25\,m) to 128 ($\geq4$\,m).
Confidence intervals come from 2,000 scene-cluster bootstrap resamples over the 33 scenes.
\cref{tab:viewpoint-bins} lists the values plotted in \cref{fig:analysis}.
3AM keeps 65\% of its near-view Recall beyond $120^\circ$ of rotation and 51\% beyond 4\,m of translation, versus 16\% and 8\% for SAM2; its gain over SAM2 grows from 15 to 40 points over rotation and from 7 to 42 points over translation, and within objects the far-minus-near gain is 18.6 points for rotation ($\geq60^\circ$ versus $<30^\circ$) and 31.6 points for translation ($\geq1$\,m versus $<0.25$\,m), with confidence intervals excluding zero.

\begin{table}[h]
\centering
\caption{\textbf{Tracking Recall by camera-pose change} on the ScanNet++ Selected Subset (\%), with the gain of 3AM over SAM2 in points and its scene-bootstrap 95\% confidence interval.}
\label{tab:viewpoint-bins}
\small
\resizebox{0.9\textwidth}{!}{
\begin{tabular}{l c cc c}
\toprule
Bin & Objects & SAM2 & 3AM & Gain [95\% CI] \\
\midrule
\multicolumn{5}{l}{\emph{Camera rotation from the prompt frame}} \\
0--15$^\circ$    & 286 & 62.2 & 77.5 & +15.2 [12.6, 18.1] \\
15--30$^\circ$   & 285 & 48.5 & 71.6 & +23.0 [20.2, 26.1] \\
30--60$^\circ$   & 286 & 36.1 & 66.0 & +29.9 [27.1, 32.3] \\
60--90$^\circ$   & 284 & 22.9 & 59.0 & +36.2 [31.8, 39.6] \\
90--120$^\circ$  & 265 & 13.4 & 53.8 & +40.5 [35.3, 44.6] \\
120--180$^\circ$ & 199 & 9.7  & 50.2 & +40.4 [32.9, 46.4] \\
\midrule
\multicolumn{5}{l}{\emph{Camera translation from the prompt frame}} \\
0--0.25\,m  & 285 & 79.2 & 86.4 & +7.3 [6.3, 8.2] \\
0.25--0.5\,m & 278 & 65.6 & 77.3 & +11.7 [9.5, 13.7] \\
0.5--1\,m   & 282 & 54.5 & 72.0 & +17.6 [14.1, 20.7] \\
1--2\,m     & 284 & 32.4 & 64.4 & +32.0 [26.3, 36.7] \\
2--4\,m     & 278 & 12.1 & 54.5 & +42.5 [37.7, 46.2] \\
$\geq4$\,m  & 128 & 6.0  & 44.4 & +38.3 [29.3, 46.9] \\
\midrule
\multicolumn{5}{l}{\emph{Within-object far-minus-near gain}} \\
Rotation, $\geq60^\circ$ vs.\ $<30^\circ$   & 285 & \multicolumn{2}{c}{near +20.1, far +38.7} & +18.6 [14.0, 23.0] \\
Translation, $\geq1$\,m vs.\ $<0.25$\,m     & 284 & \multicolumn{2}{c}{near +7.3, far +38.9}  & +31.6 [26.6, 35.6] \\
\bottomrule
\end{tabular}
}
\end{table}

\subsection{Baseline Protocols}
\label{app:eval-baselines}
\textbf{CoTracker3+SAM2 and OmniMotion+SAM2.}
Query points are sampled inside the prompt mask and tracked through the video; in every frame, the tracked points predicted as visible are given to SAM2 as point prompts, and the resulting mask is scored.
OmniMotion is optimized per video before tracking.
\textbf{PanSt3R.}
The sequence is processed offline; the predicted panoptic instance with the largest overlap with the prompt mask is used as the track.
\textbf{SegMASt3R.}
For every frame, segments are matched to the prompt frame by two-view matching, and the segment matched to the prompt mask is the prediction.
\textbf{EmbodiedSAM.}
Online 3D instance segmentation is run with the ground-truth depth and poses; the 3D instance overlapping the prompt is projected back into every frame.
\textbf{MUSt3R-Thr+SAM2.}
The cosine similarity between the MUSt3R features of the current frame and the mean feature of the prompt mask is thresholded and passed to SAM2 as a mask prompt.
\textbf{MUSt3R-Reproj+SAM2.}
The prompt mask is reprojected into every frame through MUSt3R's point maps with a visibility check, and the reprojected mask prompts SAM2.
\textbf{SAM~3.}
SAM~3 is prompted with the same mask on the prompt frame as SAM2.
\textbf{Unstructured collections.}
When the input is an unordered photo collection, frames are reordered by SAM2 feature retrieval before tracking.

\subsection{Ego-Exo4D}
\label{app:eval-egoexo}
We follow the correspondence benchmark of Ego-Exo4D~\citep{grauman2024egoexo4d} as used by V$^2$-SAM~\citep{pan2026v2sam}.
Each test sample consists of a query-view image with the ground-truth mask of one object and the synchronized target-view image.
3AM takes the query view as the prompted frame with the mask prompt and predicts the mask in the target view as the second frame; MUSt3R is initialized with the two images.
Ego frames are undistorted to a pinhole model with the nominal Aria calibration.

\subsection{Dynamic Benchmarks}
\label{app:eval-dynamic}
DAVIS17, LaSOT, and LaSOT$_{\text{ext}}$ are evaluated with their official protocols and the prompts used by DAM4SAM~\citep{videnovic2025distractor}.
For KITTI-MOTS~\citep{voigtlaender2019mots} and EPIC-KITCHENS~\citep{darkhalil2022visor}, each annotated object is prompted with its mask on the first annotated frame and tracked through the sequence; we report IoU, Tracking Recall, and Accuracy as defined in \cref{sec:setup}.

\section{Additional Results}
\label{app:results}

\paragraph{Routing Analysis.}
\label{app:routing-analysis}
For every GT-positive, non-prompt frame we measure the magnitude of each routing coefficient, namely the channel-wise RMS of $\gamma_0$ for the SAM2 term and of $\alpha_k\gamma_k$ for each geometric term $k$, where $\gamma$ includes its identity offset.
Magnitudes are normalized to shares within a frame, averaged over frames within an object, and then over objects.
These are shares of coefficient magnitudes; they exclude the shifts $\beta_k$ and the feature amplitudes, and they do not measure the contribution of each term to the fused feature.
\cref{tab:routing-branches} lists the per-term shares behind \cref{tab:routing-shares}.
Decoder blocks 4 and 11 carry the largest geometric shares on static data and almost none on dynamic data, and their signed gains ($\alpha_k\gamma_k$, mean $\pm$ SD over objects) change accordingly, block 4 from $0.48\pm0.03$ to $-0.04\pm0.04$ and block 11 from $0.47\pm0.03$ to $0.03\pm0.02$, whereas the encoder output keeps a similar gain on both ($0.39\pm0.05$ and $0.42\pm0.10$).
Decoder block 7 and the explicit geometry receive similar shares on static and dynamic data (8.7 and 10.8\%, 0.3 and 0.5\%); even so, we found that removing these two sources lowers performance, so both are kept.

\begin{table}[h]
\centering
\caption{\textbf{Per-term routing shares} (\%) on the ScanNet++ Selected Subset (static, 286 objects) and DAVIS17 (dynamic, 61 objects), with scene-bootstrap 95\% confidence intervals for the two groups.}
\label{tab:routing-branches}
\small
\resizebox{0.8\textwidth}{!}{
\begin{tabular}{l cc}
\toprule
Term & Static & Dynamic \\
\midrule
SAM2 feature ($\gamma_0$)                & 32.1 & 54.3 \\
MUSt3R encoder output                    & 17.2 & 28.0 \\
MUSt3R decoder block 4                   & 20.9 & 4.2 \\
MUSt3R decoder block 7                   & 8.7  & 10.8 \\
MUSt3R decoder block 11                  & 20.8 & 2.4 \\
Explicit geometry (point map and rays)   & 0.3  & 0.5 \\
\midrule
Appearance terms (SAM2, encoder)         & 49.3 & 82.2 \\
Geometric terms (blocks 4, 7, 11, explicit) & 50.7 [50.4, 51.0] & 17.8 [16.8, 18.9] \\
\bottomrule
\end{tabular}
}
\end{table}

\paragraph{Memory Selection.}
3AM uses SAM2's original memory selection.
\cref{tab:memory-selection} combines 3AM with the memory-selection policies of DAM4SAM and SAM2Long on the ScanNet++ Selected Subset.
Both policies change the results by less than one point in either direction, far less than the gain of 3AM over the same policies applied to SAM2 (\cref{tab:main}), which shows that the improvement comes from the features rather than from memory management.
Memory selection tailored to geometry-aware memories is an interesting direction for future work.

\begin{table}[h]
\centering
\caption{\textbf{Memory-selection policies on top of 3AM}, ScanNet++ Selected Subset (\%).}
\label{tab:memory-selection}
\small
\begin{tabular}{lccc}
\toprule
Method & IoU & Recall & Acc. \\
\midrule
3AM (SAM2 memory selection)                          & 90.6 & 65.7 & 72.4 \\
3AM + DAM4SAM~\citep{videnovic2025distractor} policy & 90.4 & 66.4 & 72.0 \\
3AM + SAM2Long~\citep{ding2024sam2long} policy       & 90.8 & 65.1 & 73.6 \\
\bottomrule
\end{tabular}
\end{table}

\paragraph{Prompt Type and Unordered Input.}
\cref{tab:input-modality} evaluates mask, point, box, and point-plus-box prompts, and unordered photo collections, on the Selected Subset.
3AM keeps its advantage for every prompt type, and the drop on unordered collections is small, because MUSt3R's memory does not require temporal continuity.

\begin{table}[h]
\centering
\caption{\textbf{Prompt types and unordered input}, ScanNet++ Selected Subset (IoU / Tracking Recall / Accuracy, \%).
For unordered photo collections the frames are shuffled and reordered by SAM2 feature retrieval before tracking.}
\label{tab:input-modality}
\vspace{-2mm}
\small
\begin{tabular}{lccc}
\toprule
Prompt / input & SAM2 & DAM4SAM & 3AM \\
\midrule
Mask                          & 74.3 / 28.2 / 66.6 & 76.6 / 37.5 / 63.8 & 90.6 / 65.7 / 72.4 \\
Mask, unordered collection    & -- & 71.7 / 32.1 / 58.6 & 90.1 / 65.2 / 72.1 \\
Point                         & -- & 75.1 / 35.1 / 61.4 & 88.7 / 64.2 / 71.4 \\
Box                           & -- & 74.3 / 34.7 / 60.7 & 87.6 / 63.3 / 70.9 \\
Point + box                   & -- & 74.6 / 34.4 / 61.1 & 88.0 / 62.6 / 71.0 \\
\bottomrule
\end{tabular}
\end{table}

\paragraph{Repeated Trajectories.}
To test long-range stability, each Selected Subset sequence is repeated one, two, or four times so that the object reappears again and again from the same viewpoints.
\cref{tab:cycles} shows that 3AM stays stable across cycles and DAM4SAM does not improve with repetition, whereas SAM2 gains from the repeated views (39.0 Recall at two cycles) but remains far below 3AM.

\begin{table}[h]
\centering
\caption{\textbf{Repeated trajectories}, ScanNet++ Selected Subset (IoU / Tracking Recall / Accuracy, \%).}
\label{tab:cycles}
\small
\begin{tabular}{cccc}
\toprule
Cycles & SAM2 & DAM4SAM & 3AM \\
\midrule
1 & 74.3 / 28.2 / 66.6 & 76.6 / 37.5 / 63.8 & 90.6 / 65.7 / 72.4 \\
2 & 84.5 / 39.0 / 80.2 & 75.5 / 35.7 / 61.5 & 91.8 / 67.9 / 74.0 \\
4 & 83.2 / 36.6 / 79.5 & 73.9 / 34.4 / 61.2 & 91.4 / 67.1 / 74.5 \\
\bottomrule
\end{tabular}
\end{table}

\section{MUSt3R Compared with Other Reconstruction Backbones as Feature Sources}
\label{app:feature-analysis}
\cref{tab:3D_backbone} compares reconstruction backbones as the feature source of 3AM (\cref{sec:ablation}); VGGT and $\pi^3$ cannot process the evaluation sequences online, and CUT3R features give low Recall.

\begin{table}[h]
\centering
\caption{\textbf{3D reconstruction backbones} as the feature source of 3AM. Frame Num.: frames processable in one sequence; Tracking Recall on the ScanNet++ Selected Subset. Best in red.}
\label{tab:3D_backbone}
\small
\begin{tabular}{l ccc}
\toprule
Backbone & Online & Frame Num. & Recall \\
\midrule
VGGT~\citep{wang2025vggt}        & \xmark & $\sim$200   & -- \\
$\pi^3$~\citep{wang2025pi3}      & \xmark & $\sim$300   & -- \\
CUT3R~\citep{wang2025continuous} & \cmark & $>$10,000 & 27.5 \\
MUSt3R~\citep{cabon2025must3r}   & \cmark & $>$10,000 & \best{65.7} \\
\bottomrule
\end{tabular}
\end{table}

\cref{sec:feature-analysis} analyzes MUSt3R across layers.
Here we ask whether another reconstruction backbone would be a better feature source, using the frozen features of each model without fine-tuning, which is how 3AM consumes them.
We compare MUSt3R-512 with LingBot-Map~\citep{chen2026lingbotmap}, a recent streaming model with a compact geometric context, and with the offline VGGT-1B~\citep{wang2025vggt}, on ASE (6 scenes, 16 keyframes each), ScanNet++ (8 scenes, 16 frames within a 64-frame window), and MOSE (8 dynamic videos, 12 frames each).
Online models stream over the same capped window; VGGT sees the 16 evaluation frames jointly.
Ground-truth correspondences are obtained by unprojecting the depth of one frame into the other with a depth-consistency check.
For every model we report its best layer.
\cref{tab:featcorr} summarizes the results.
MUSt3R gives the best dense correspondence (PCK and retrieval accuracy) on both static datasets and the most peaked similarity maps (lowest entropy) on ScanNet++, better than VGGT although VGGT sees all evaluation frames jointly and cannot be run online over a long stream; on ASE, VGGT's similarity maps are more peaked (0.751 versus 0.826 entropy) while MUSt3R still matches more points correctly.
LingBot-Map's similarities are globally saturated, with cosine values above 0.9 almost everywhere and near-zero margins between the true match and the hardest negative.
On the dynamic MOSE videos, MUSt3R keeps a clear contrast between an object and its background over time (cosine 0.88 within the object versus 0.34 to the background), whereas LingBot-Map hardly separates them; this contrast is what memory attention needs to track a specific object.
VGGT is also best at instance-level discrimination on ASE and at metric geometry (Spearman correlation with 3D distance), which matches the layer-wise picture in \cref{sec:feature-analysis}, where correspondence and metric geometry are carried by different representations, and MUSt3R's middle layers give the correspondence 3AM relies on.

\begin{table}[h]
\centering
\caption{\textbf{Frozen features of reconstruction backbones as correspondence sources.}
PCK@5 is the fraction of dense matches within 5\% of the image diagonal, Ret.\ the retrieval accuracy of the true match, Ent.\ the normalized entropy of the similarity map (lower is more peaked), Inst.\ the cross-frame instance nearest-neighbour accuracy, $\rho$ the magnitude of the Spearman correlation between feature similarity and 3D distance, Prop.\ the MOSE mask-propagation mIoU by feature nearest neighbours, and Contrast the cosine gap between same-object and object-to-background similarities over time. Best values in red.}
\label{tab:featcorr}
\small
\resizebox{\textwidth}{!}{
\begin{tabular}{l c ccccc ccccc cc}
\toprule
& & \multicolumn{5}{c}{ASE} & \multicolumn{5}{c}{ScanNet++} & \multicolumn{2}{c}{MOSE} \\
\cmidrule(lr){3-7}\cmidrule(lr){8-12}\cmidrule(lr){13-14}
Backbone & Online & PCK@5 & Ret. & Ent.$\downarrow$ & Inst. & $\rho$ & PCK@5 & Ret. & Ent.$\downarrow$ & Inst. & $\rho$ & Prop. & Contrast \\
\midrule
MUSt3R~\citep{cabon2025must3r}        & \cmark & \best{45.6} & \best{46.5} & 0.826 & 50.6 & 0.648 & \best{72.4} & \best{75.3} & \best{0.807} & 73.7 & 0.777 & 54.5 & \best{0.54} \\
LingBot-Map~\citep{chen2026lingbotmap} & \cmark & 43.0 & 39.3 & 0.971 & 58.3 & 0.568 & 54.8 & 55.8 & 0.935 & 69.0 & 0.605 & 50.1 & 0.02 \\
VGGT-1B~\citep{wang2025vggt}          & \xmark & 44.3 & 40.7 & \best{0.751} & \best{64.8} & \best{0.728} & 55.4 & 56.1 & 0.839 & \best{74.6} & \best{0.838} & \best{56.8} & 0.33 \\
\bottomrule
\end{tabular}
}
\end{table}

\section{Class-Agnostic 3D Instance Segmentation}
\label{app:class-agnostic}
Given keyframes sampled at a fixed stride, we first generate 2D instance proposals with SAM2's automatic mask generator.
For each keyframe, its proposals are propagated forward with 3AM to all later frames; no backward tracking is performed.
The propagated masks are lifted into 3D by back-projecting their pixels with the ground-truth depth and poses.
Reprojection errors concentrate near object boundaries, so we compute a reprojection score from the agreement between projected and observed depth and erode the 2D masks slightly before projection.
Per-frame 3D fragments are then merged into instances with two scores, a 3D overlap score in the point-cloud domain and a 2D temporal-overlap score along the propagated tracks.
Because 3AM keeps the identity of an object across the video, an object seen earlier is recognized again when it reappears, which provides reliable temporal evidence for merging.
Duplicate assignments are resolved by majority voting at the superpoint level, so that each superpoint is assigned to the instance in which it is observed most often.

\section{Limitations}
\label{app:limitations}
An object that is moved while it is outside the field of view, for example behind the camera, cannot be recovered immediately.
Its geometric anchor points to the old location, no visual evidence links the old and the new location, and neither appearance nor geometry can re-establish the identity until the object is observed again.

\section{Additional Qualitative Results}
\label{app:qualitative}
\begin{figure}[p]
    \centering
    \includegraphics[width=\textwidth]{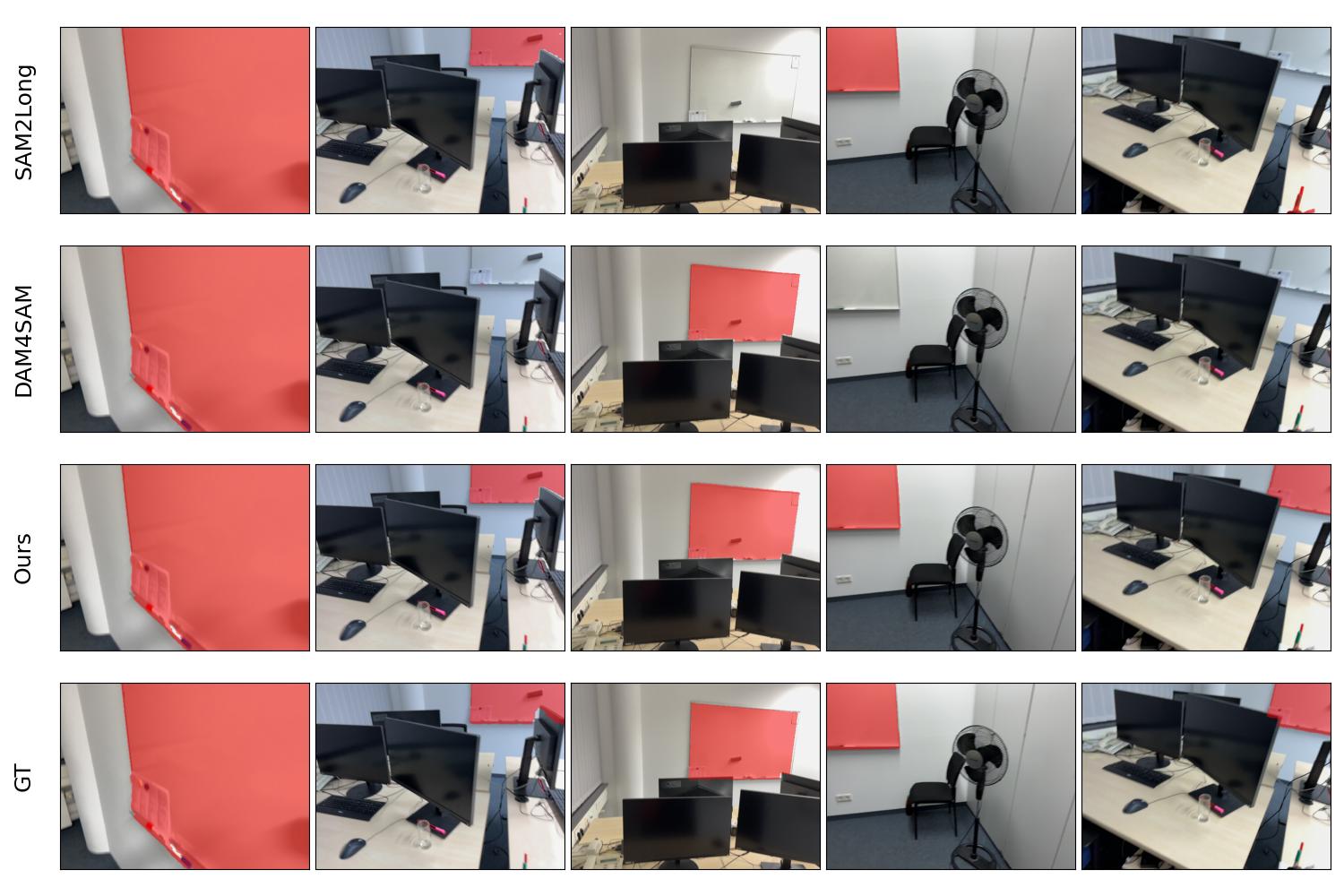}\\[2mm]
    \includegraphics[width=\textwidth]{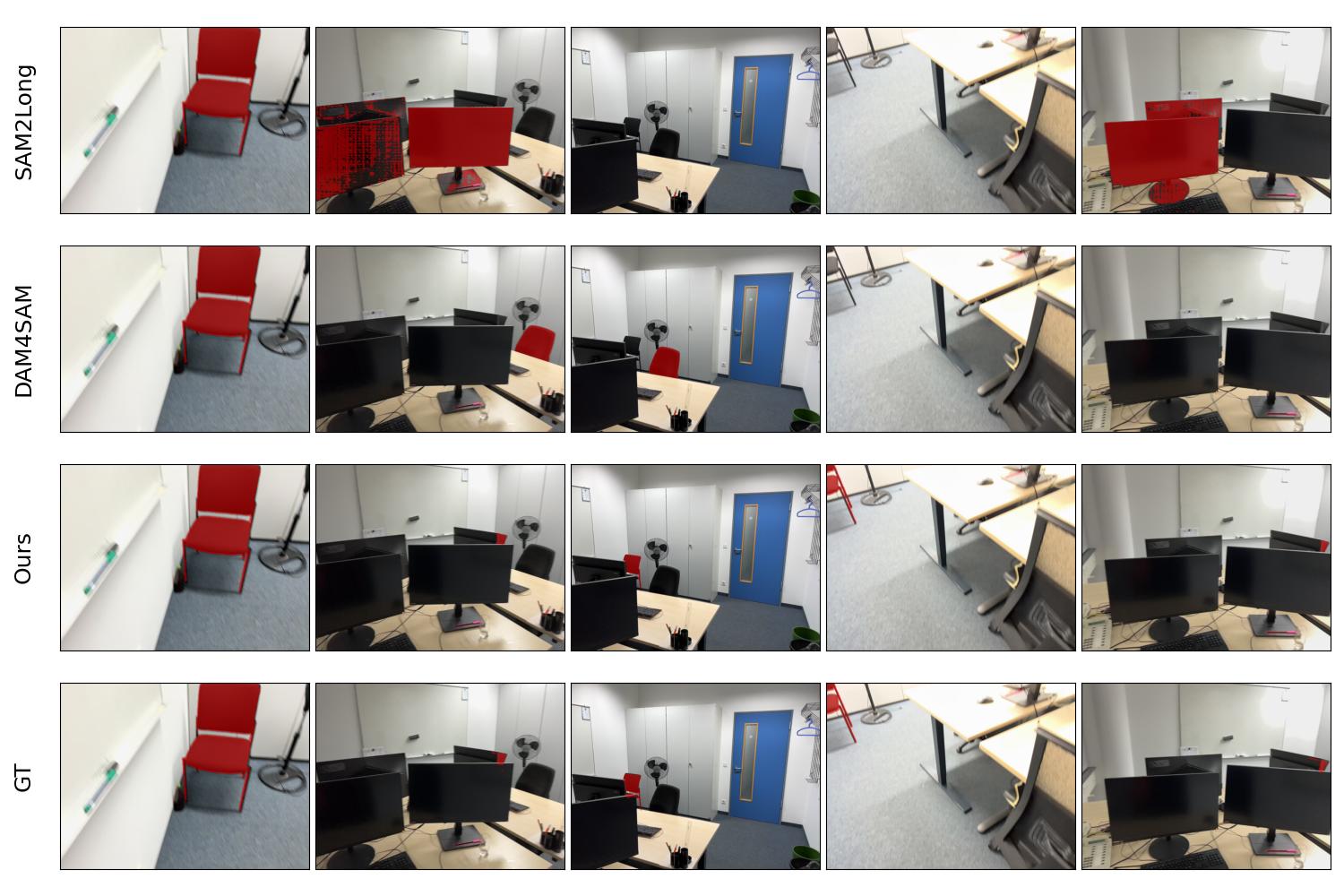}
    \caption{\textbf{Qualitative comparison of VOS methods} (scene 1ada7a0617).}
    \label{fig:app_vis1}
\end{figure}

\begin{figure}[p]
    \centering
    \includegraphics[width=\textwidth]{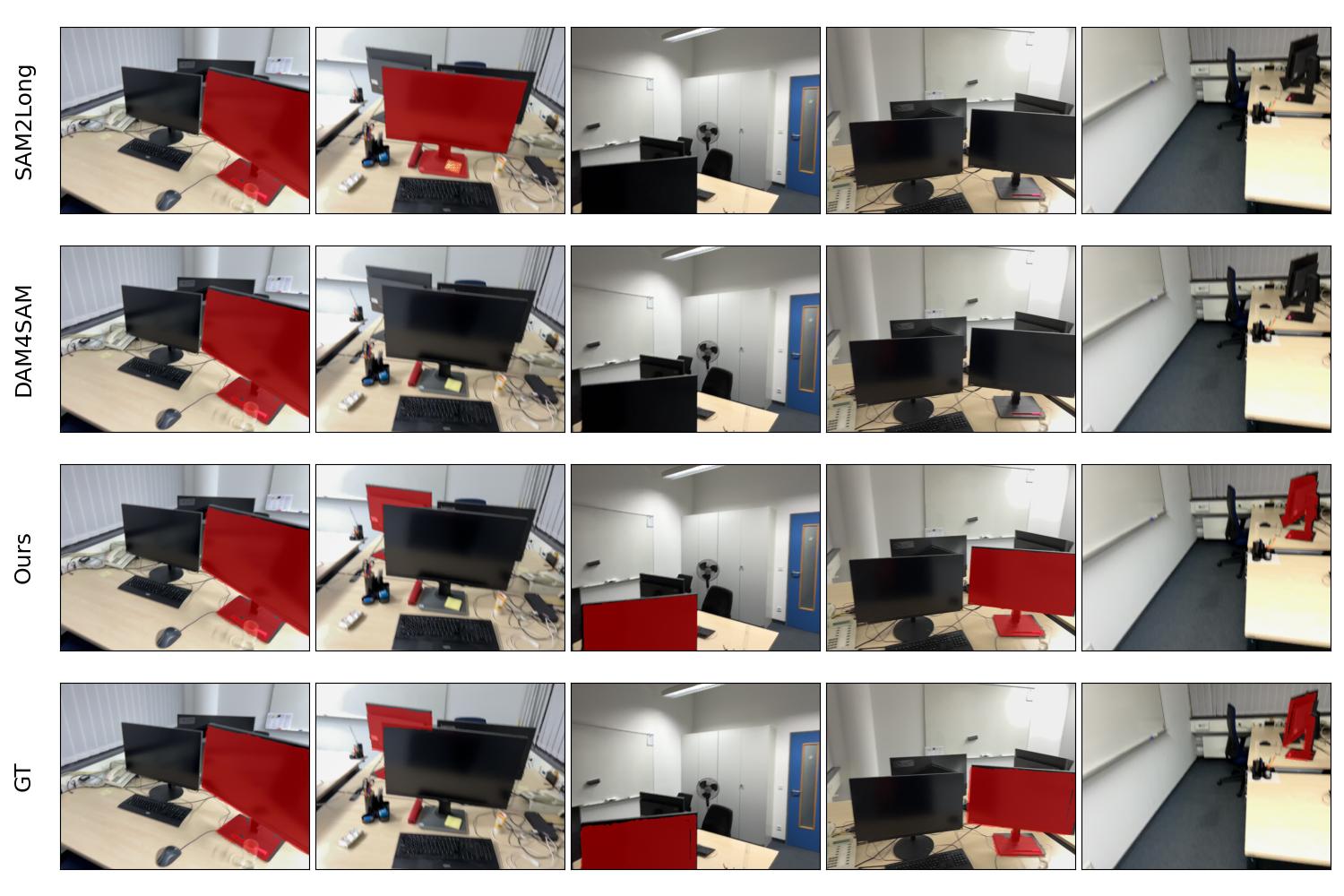}\\[2mm]
    \includegraphics[width=\textwidth]{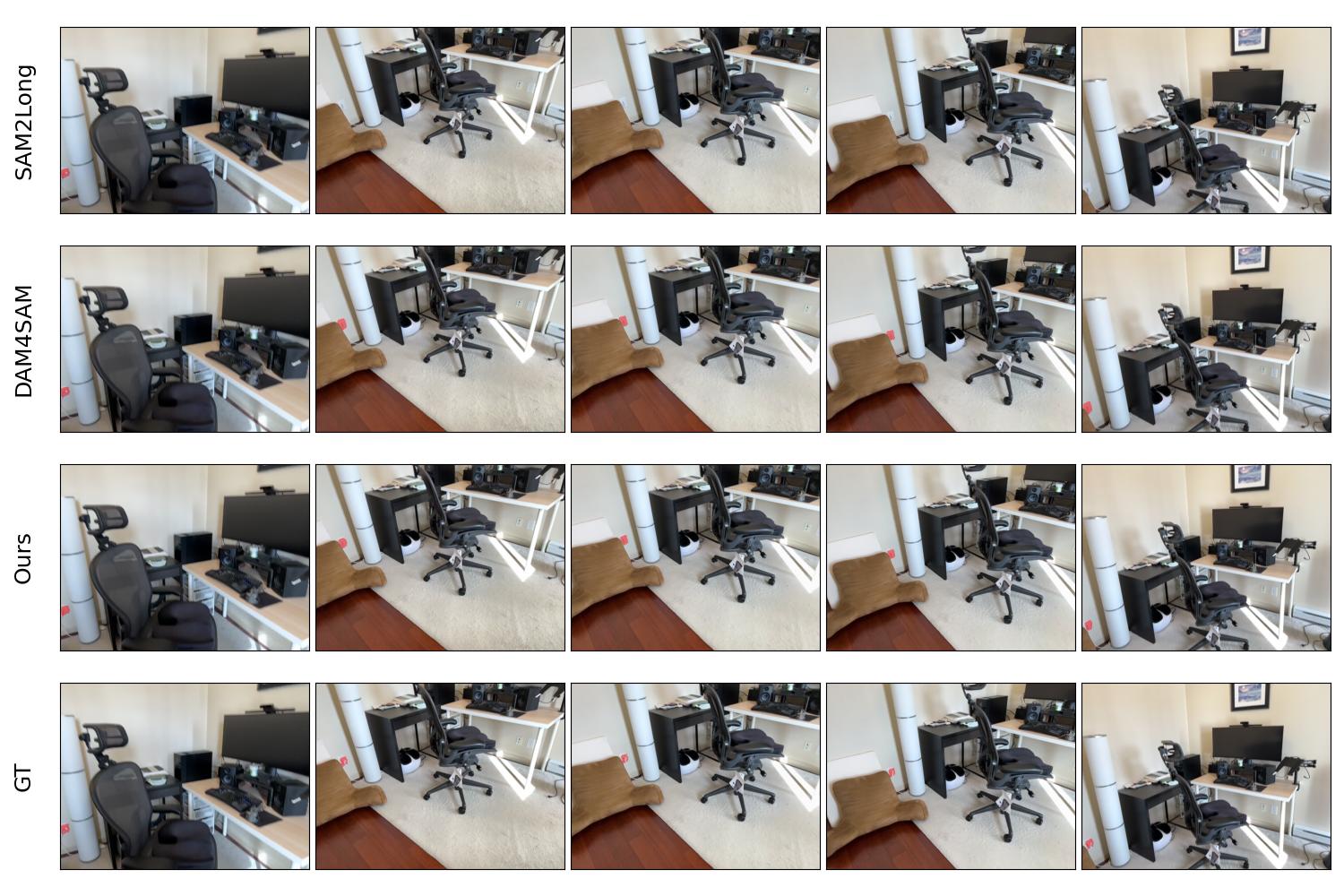}
    \caption{\textbf{Qualitative comparison of VOS methods} (scenes 1ada7a0617 and a24f64f7fb).}
    \label{fig:app_vis2}
\end{figure}

\begin{figure}[p]
    \centering
    \includegraphics[width=\textwidth]{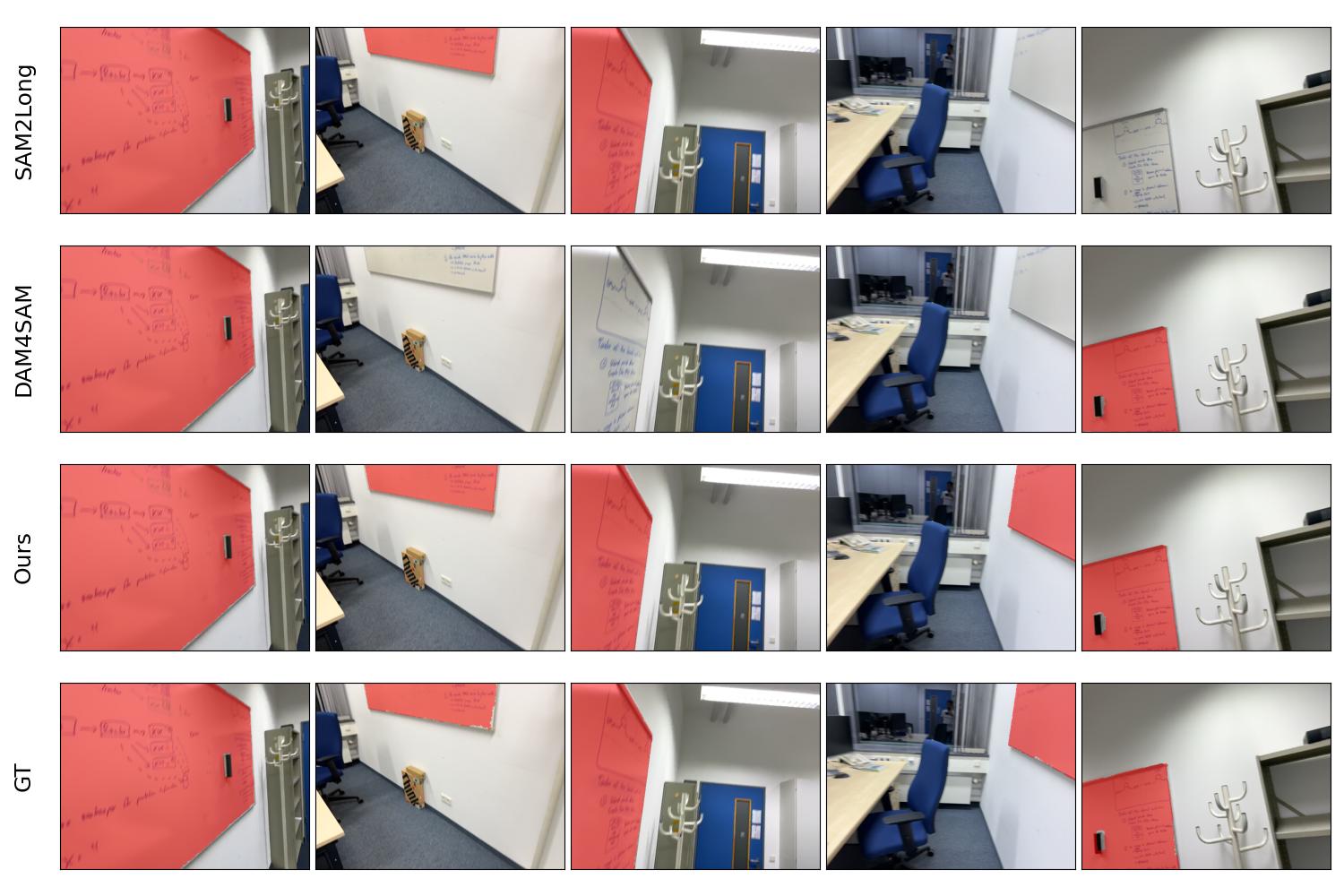}\\[2mm]
    \includegraphics[width=\textwidth]{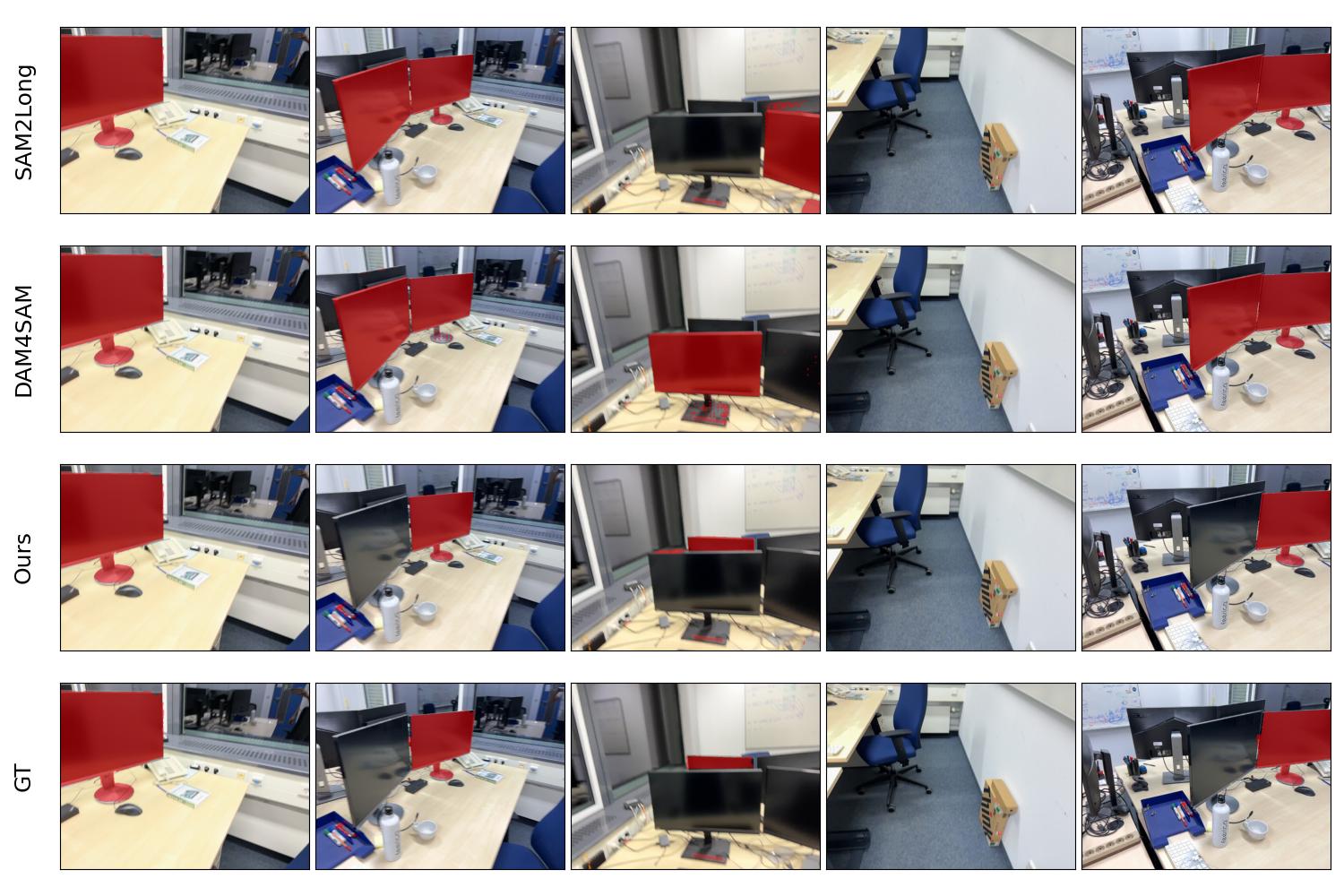}
    \caption{\textbf{Qualitative comparison of VOS methods} (scene 7b6477cb95).}
    \label{fig:app_vis3}
\end{figure}

\begin{figure}[p]
    \centering
    \includegraphics[width=\textwidth]{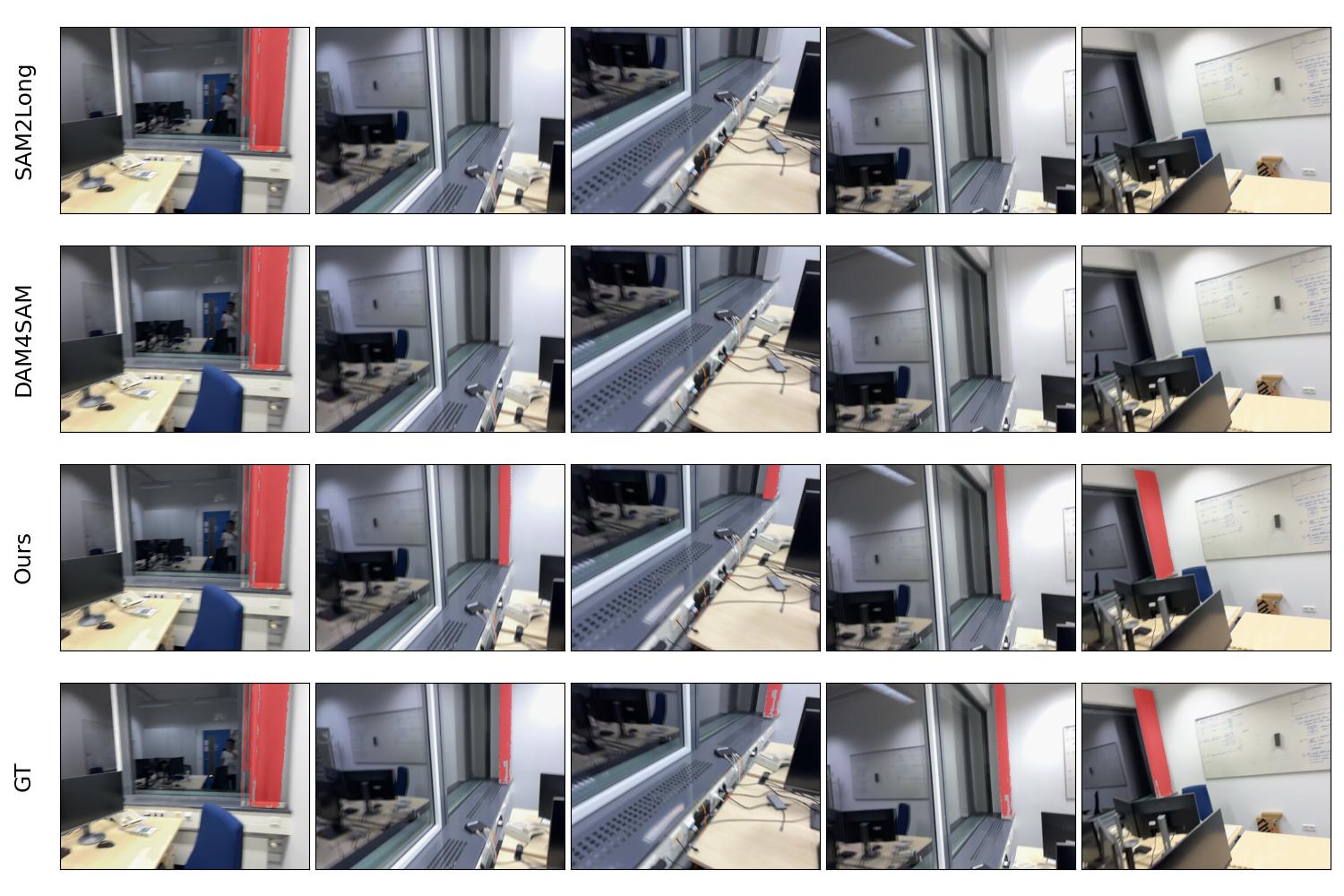}\\[2mm]
    \includegraphics[width=\textwidth]{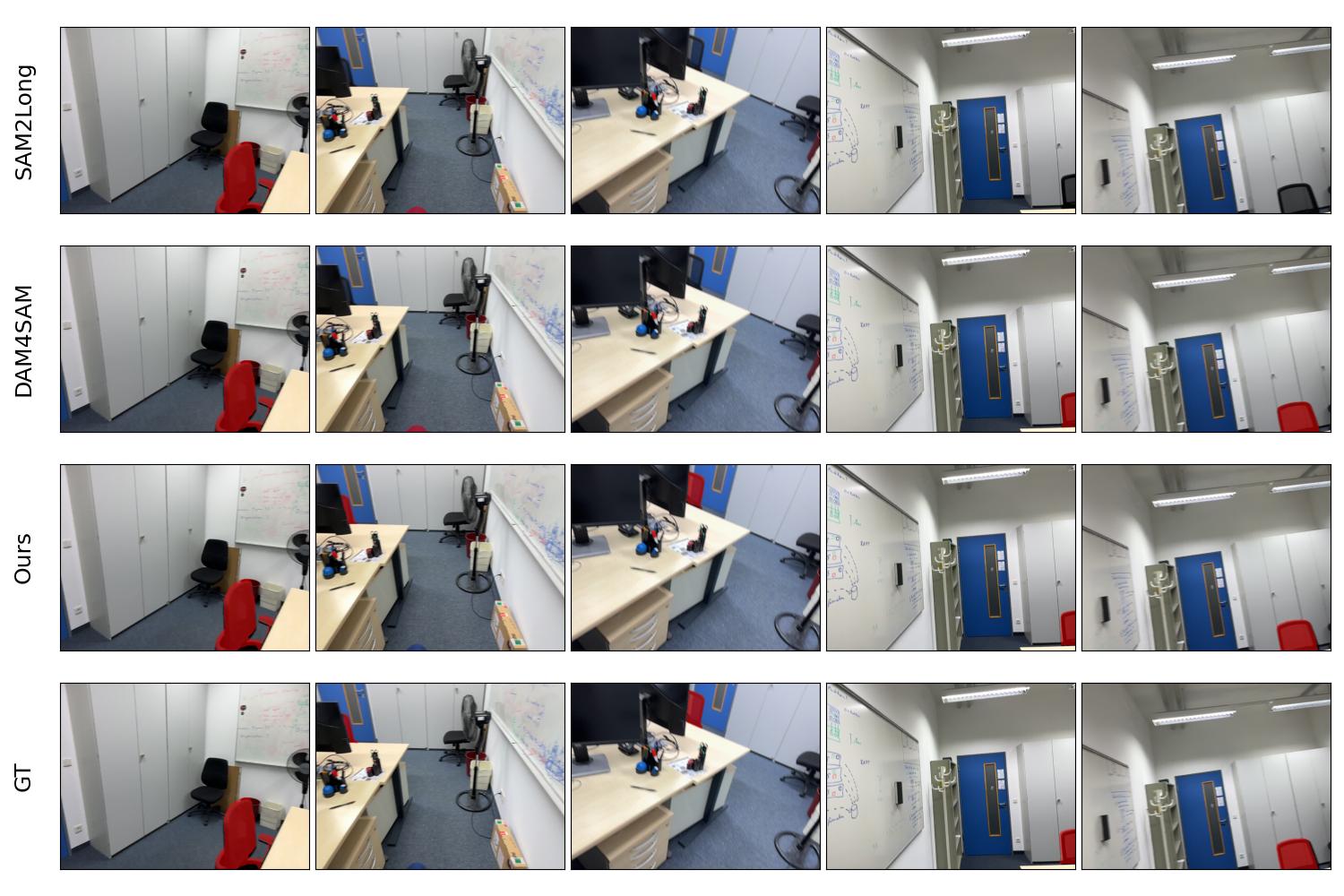}
    \caption{\textbf{Qualitative comparison of VOS methods} (scene 7b6477cb95).}
    \label{fig:app_vis4}
\end{figure}

\begin{figure}[p]
    \centering
    \includegraphics[width=\textwidth]{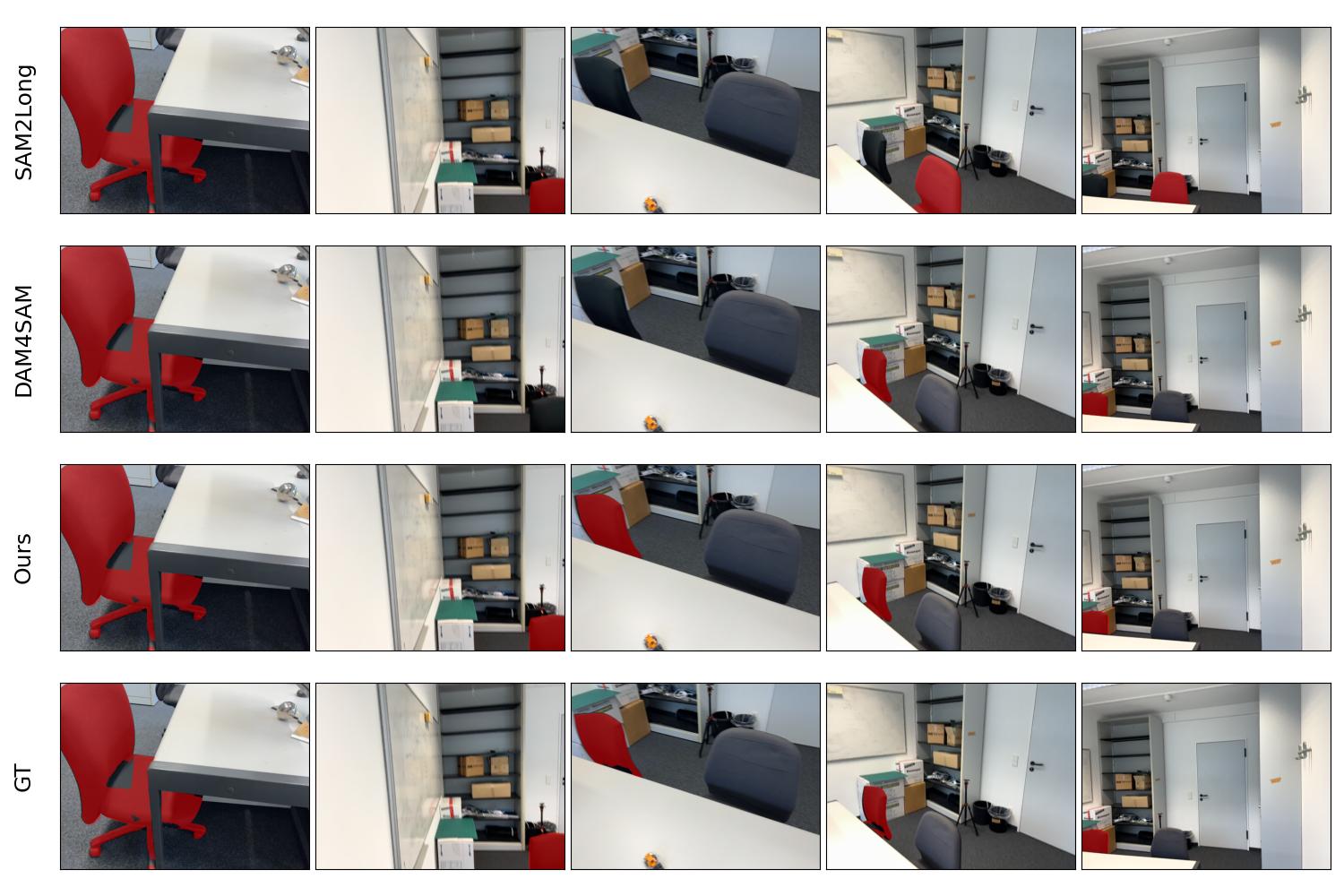}\\[2mm]
    \includegraphics[width=\textwidth]{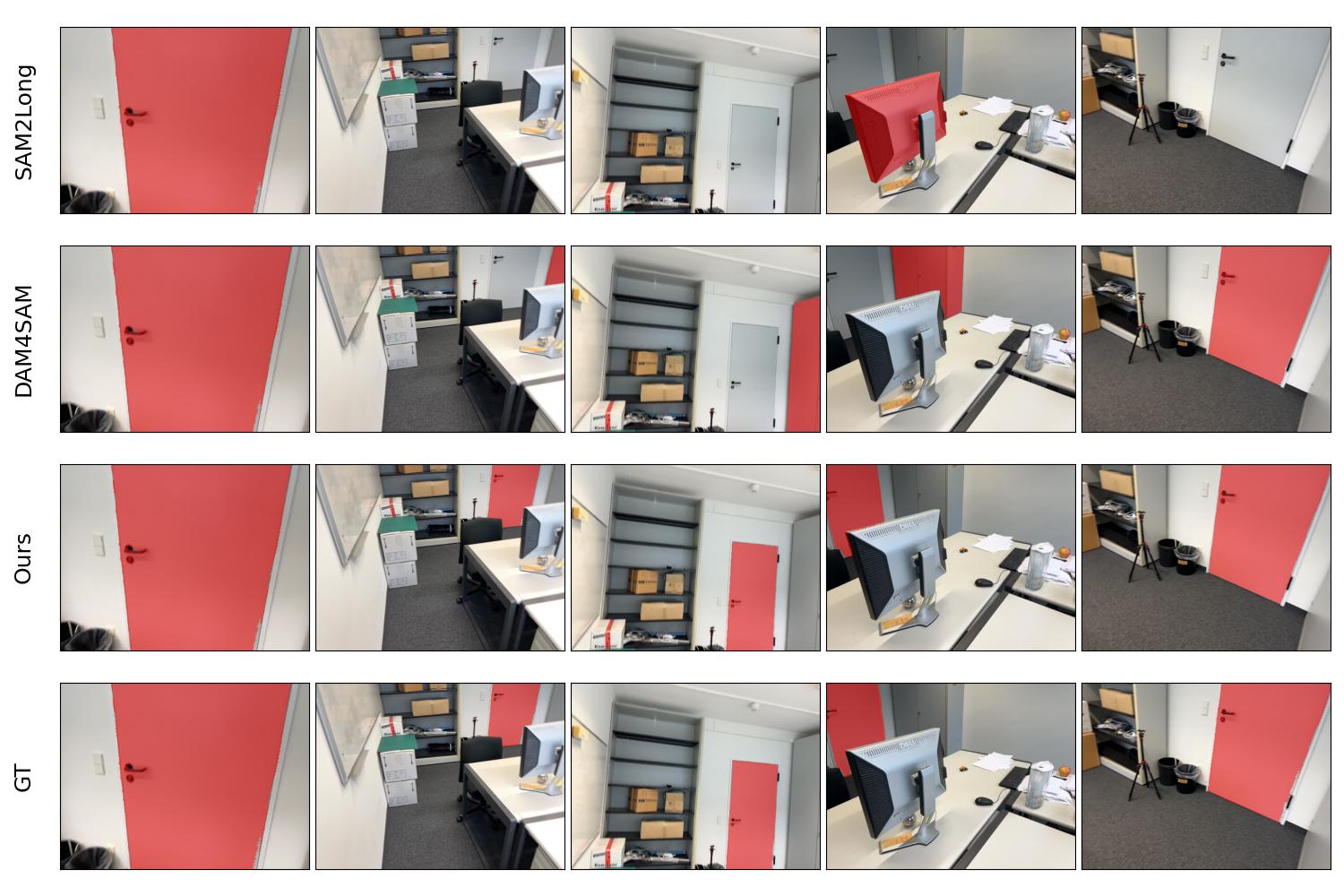}
    \caption{\textbf{Qualitative comparison of VOS methods} (scene acd95847c5).}
    \label{fig:app_vis5}
\end{figure}

\begin{figure}[p]
    \centering
    \includegraphics[width=\textwidth]{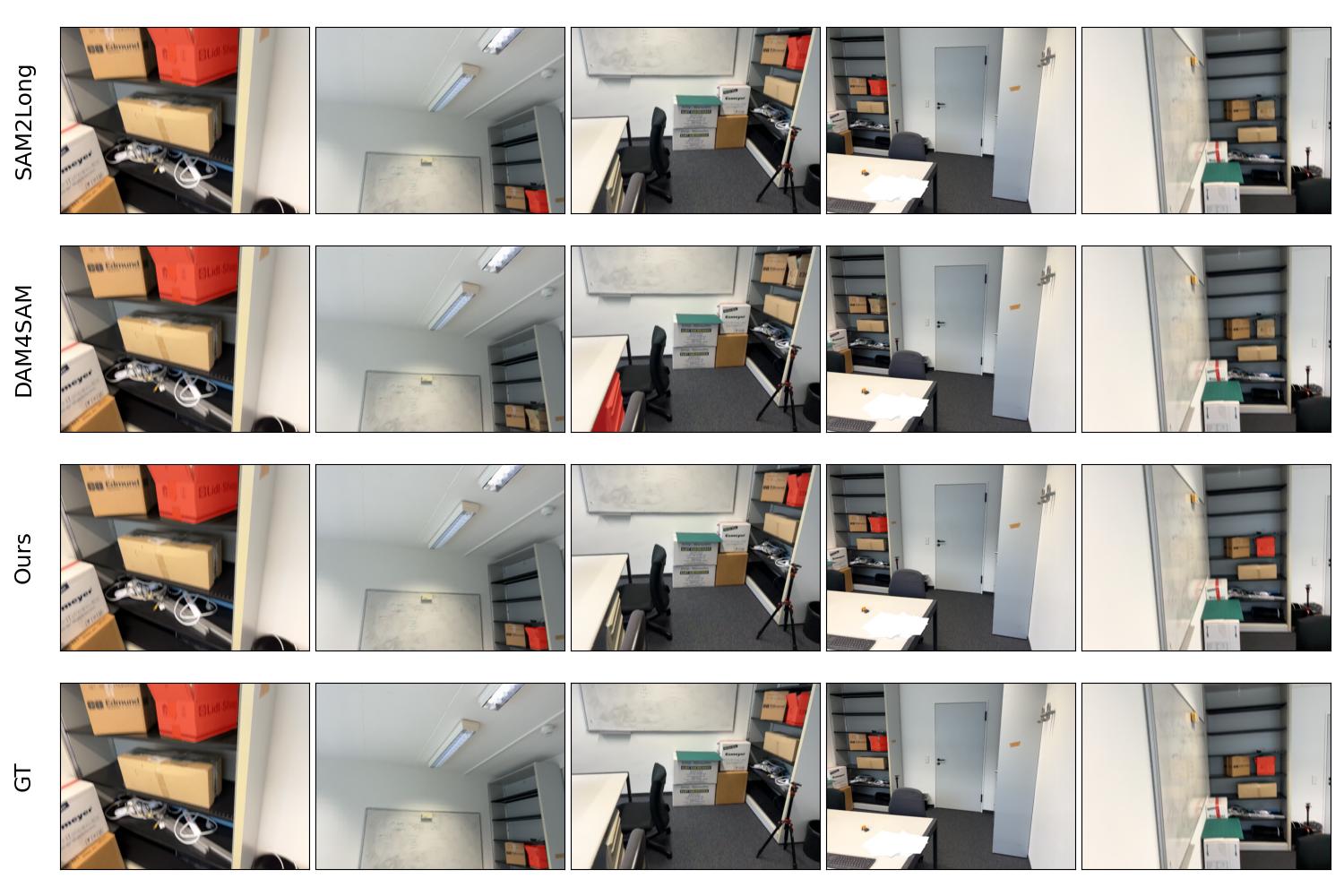}\\[2mm]
    \includegraphics[width=\textwidth]{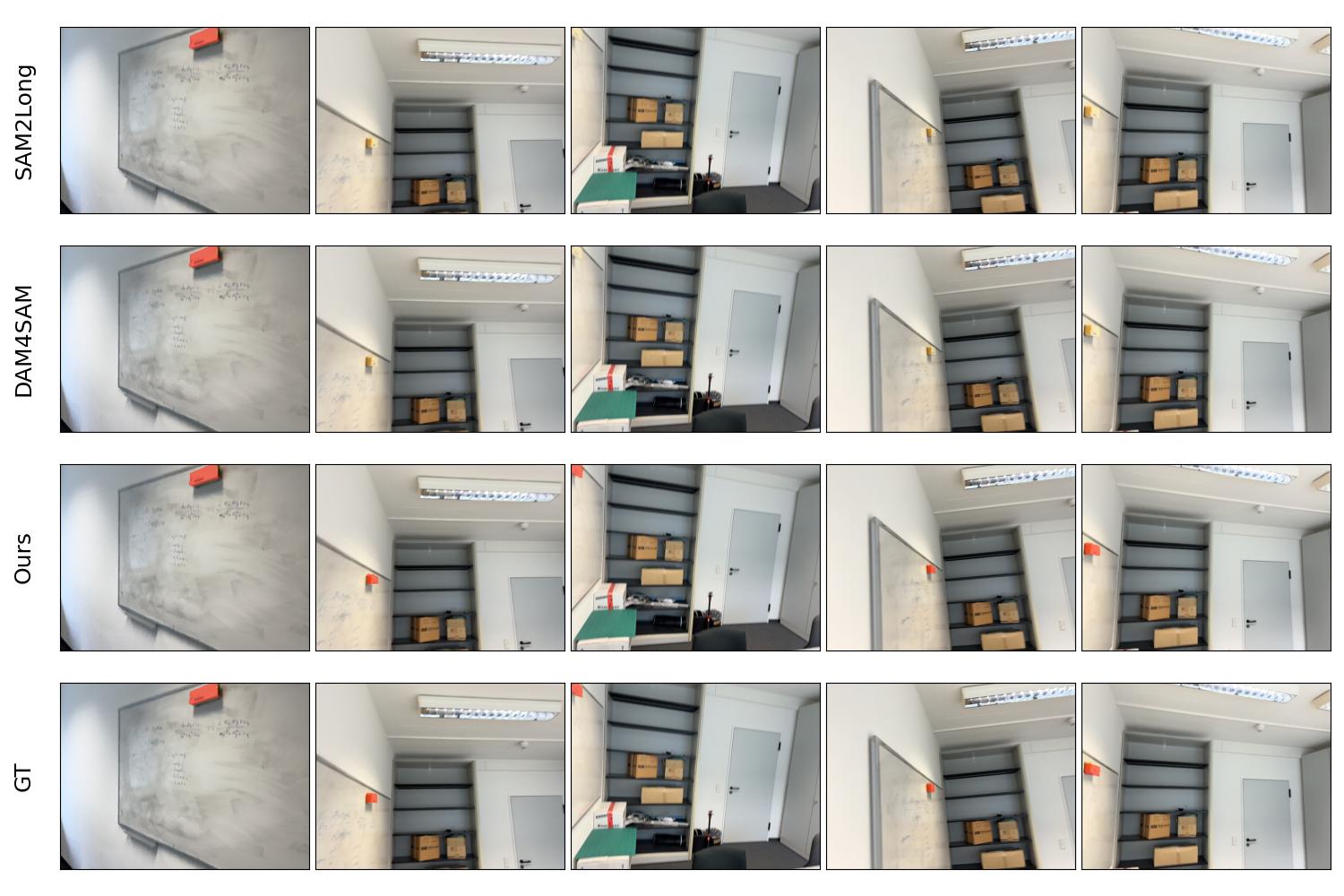}
    \caption{\textbf{Qualitative comparison of VOS methods} (scene acd95847c5).}
    \label{fig:app_vis6}
\end{figure}

\begin{figure}[p]
    \centering
    \includegraphics[width=\textwidth]{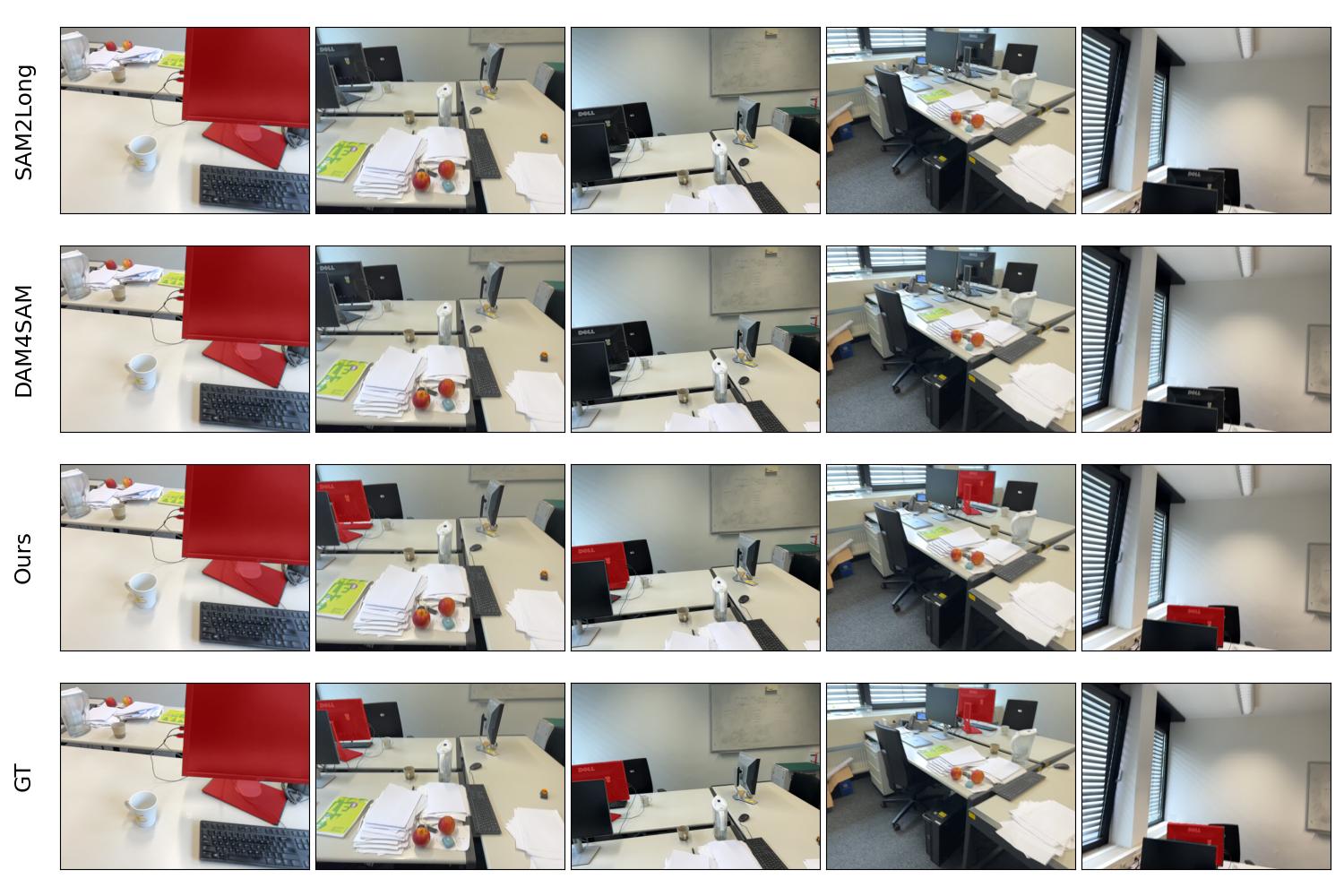}\\[2mm]
    \includegraphics[width=\textwidth]{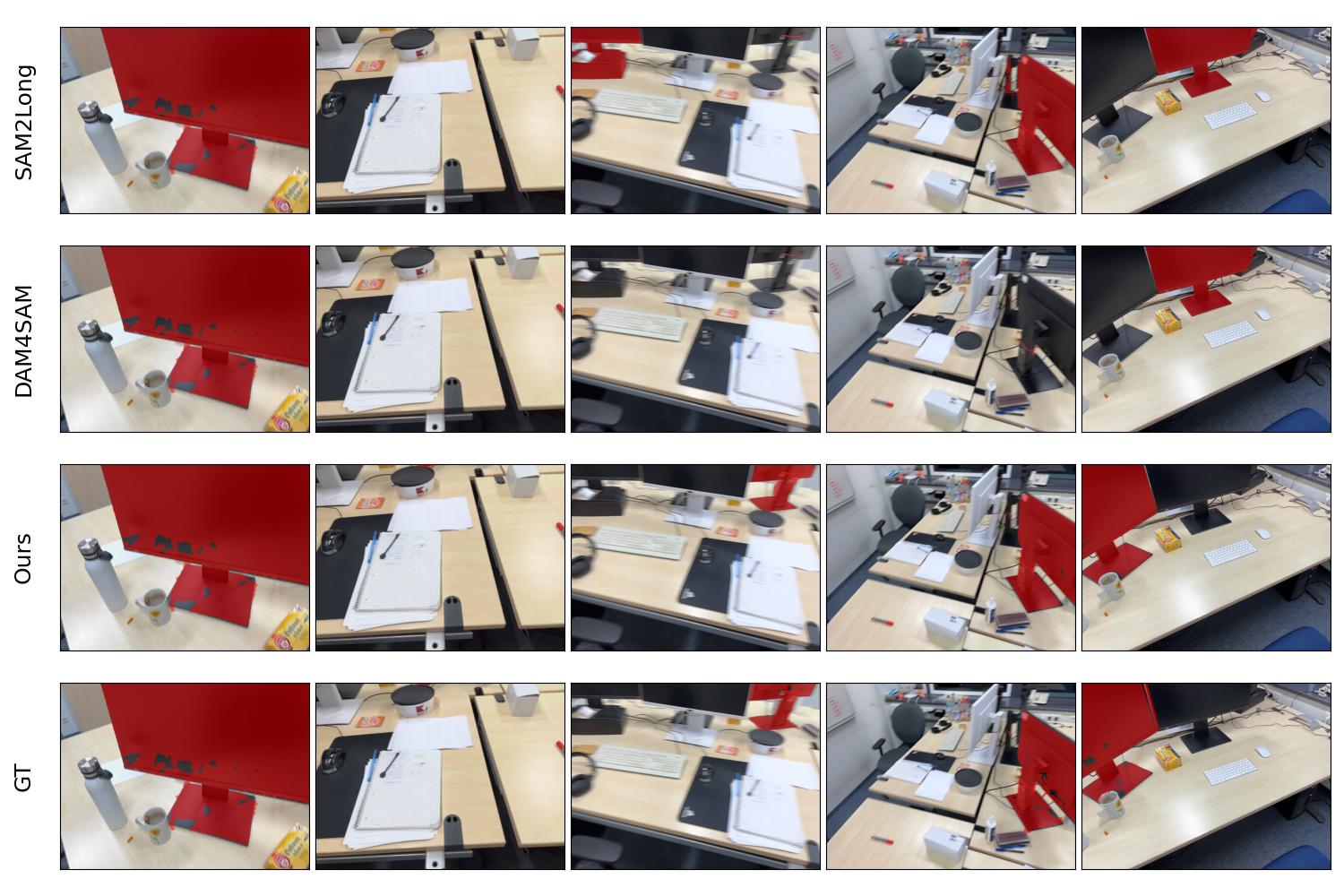}
    \caption{\textbf{Qualitative comparison of VOS methods} (scenes acd95847c5 and c50d2d1d42).}
    \label{fig:app_vis7}
\end{figure}

\begin{figure}[p]
    \centering
    \includegraphics[width=\textwidth]{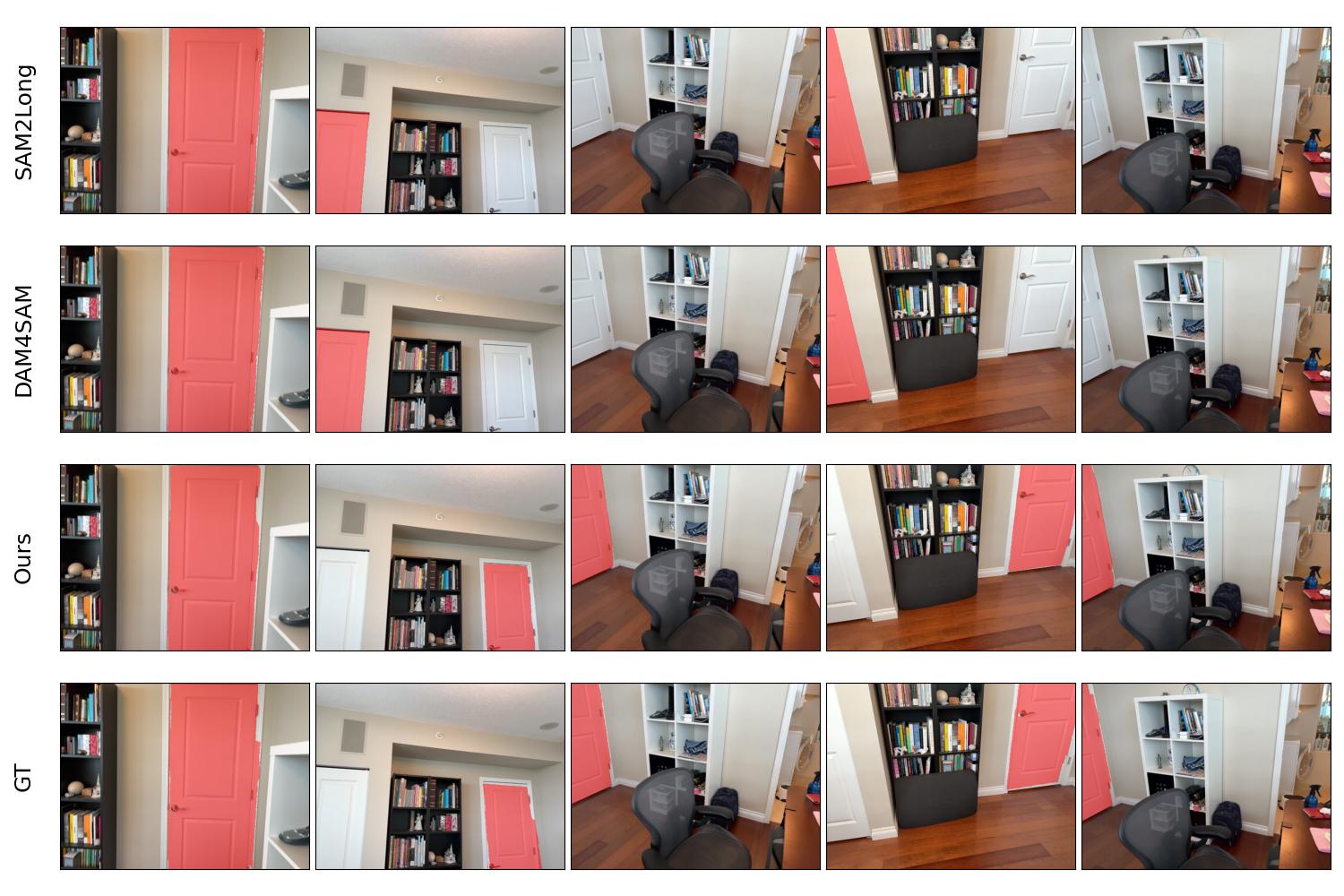}
    \caption{\textbf{Qualitative comparison of VOS methods} (scene cc5237fd77).}
    \label{fig:app_vis8}
\end{figure}

\begin{figure}[p]
\centering
\setlength{\tabcolsep}{2pt}
\begin{tabular}{ccc}
\includegraphics[width=0.32\textwidth]{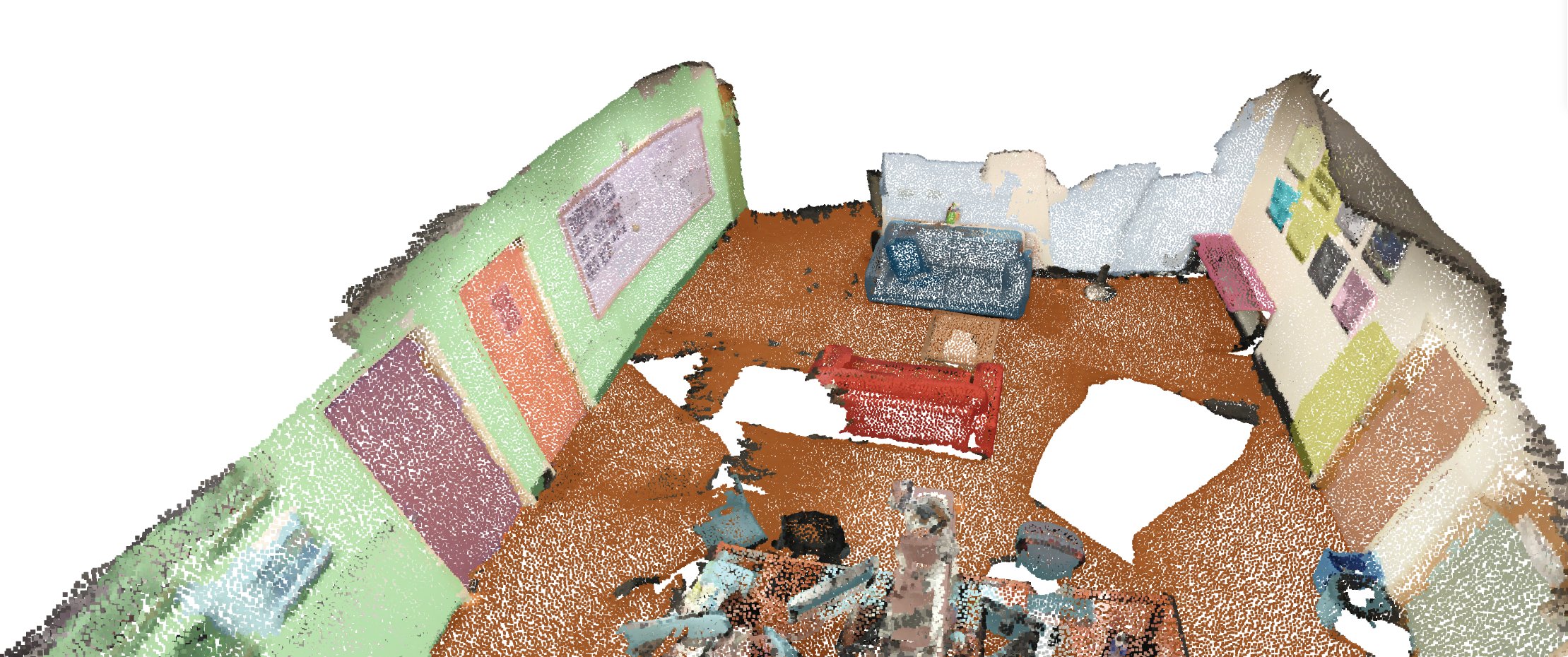} &
\includegraphics[width=0.32\textwidth]{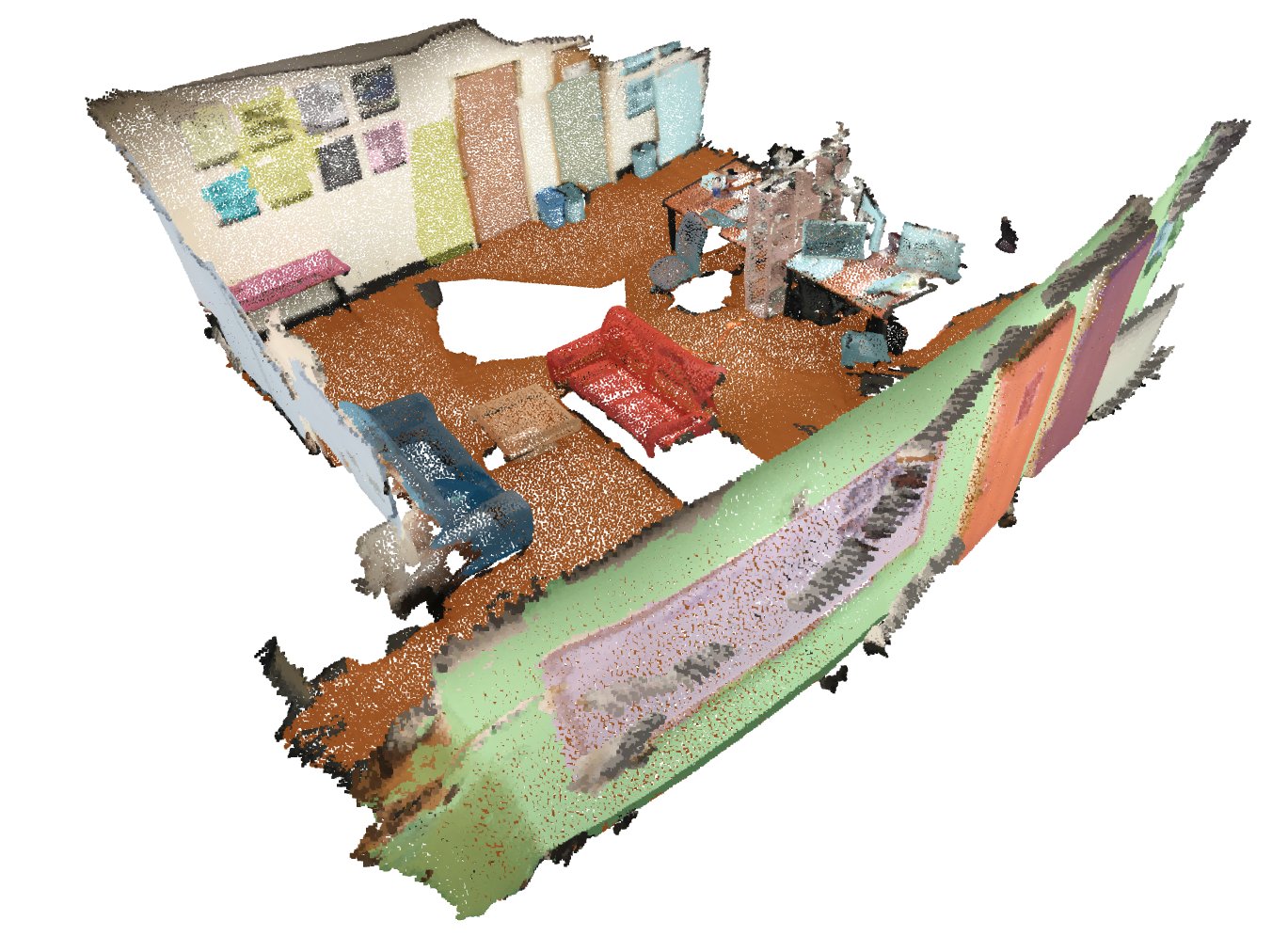} &
\includegraphics[width=0.32\textwidth]{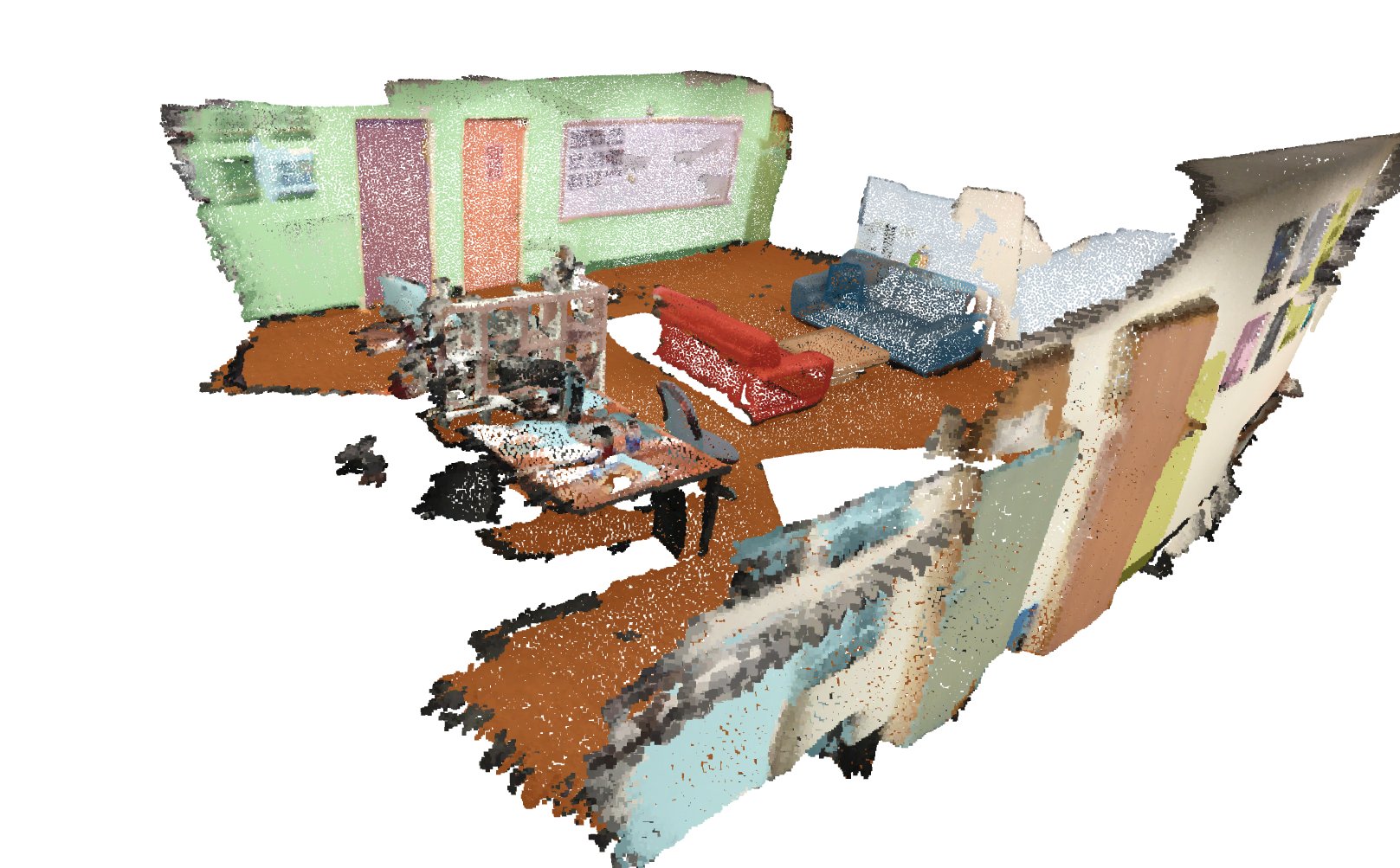} \\
\includegraphics[width=0.32\textwidth]{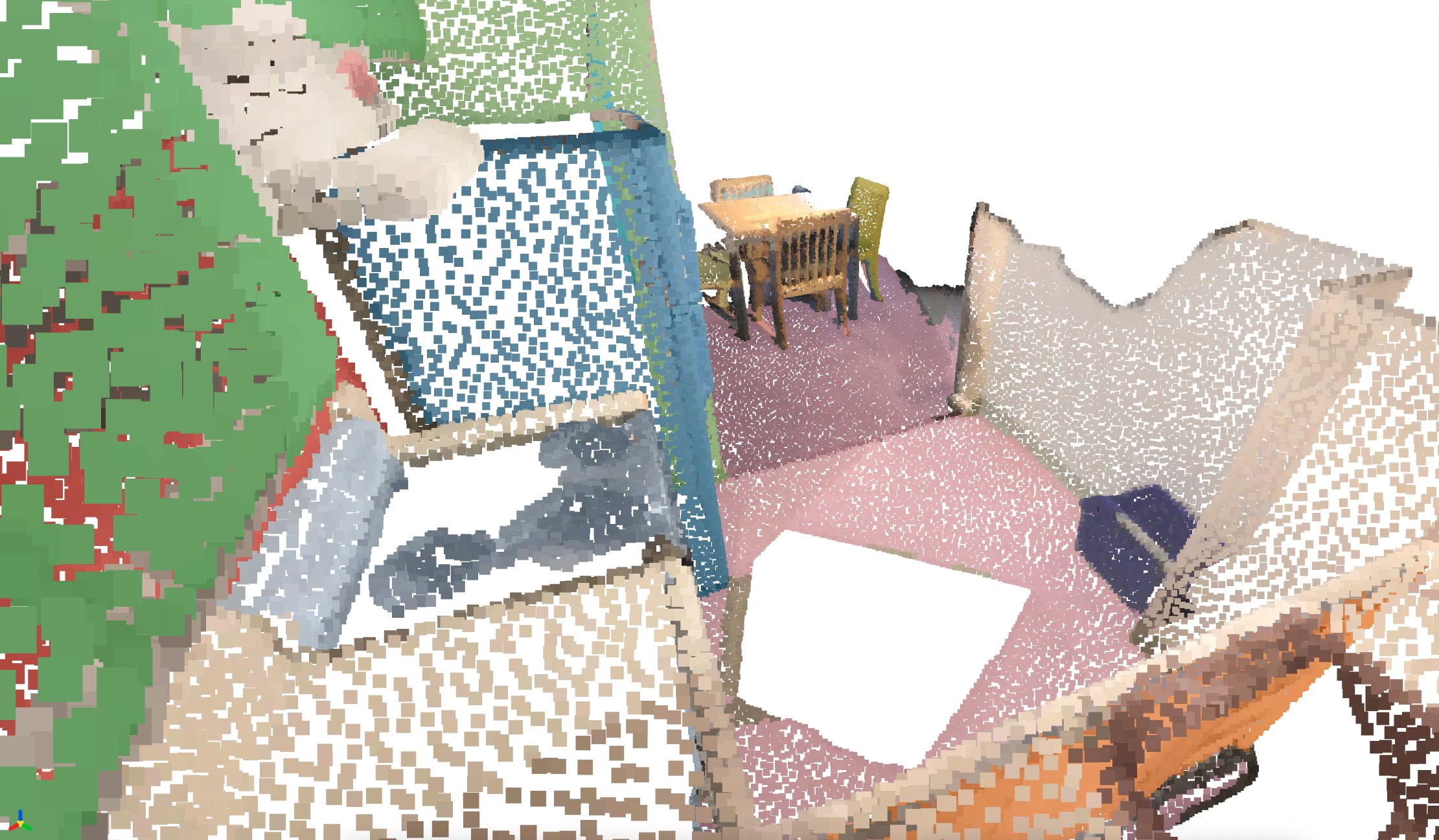} &
\includegraphics[width=0.32\textwidth]{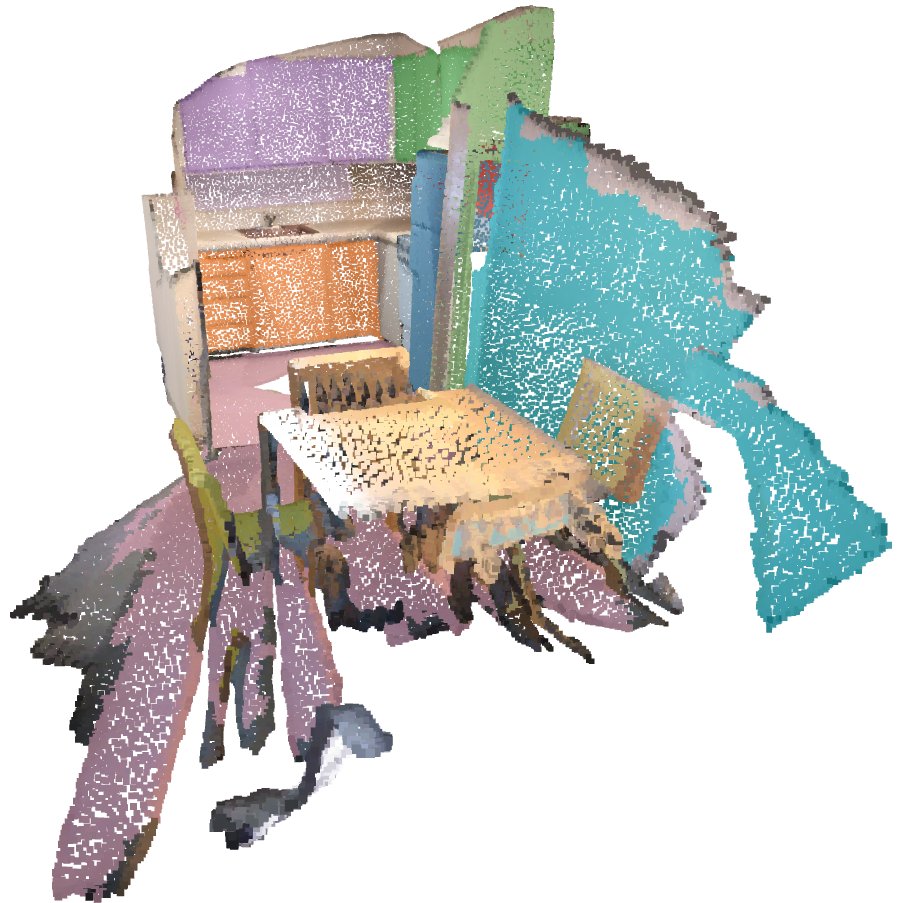} &
\includegraphics[width=0.32\textwidth]{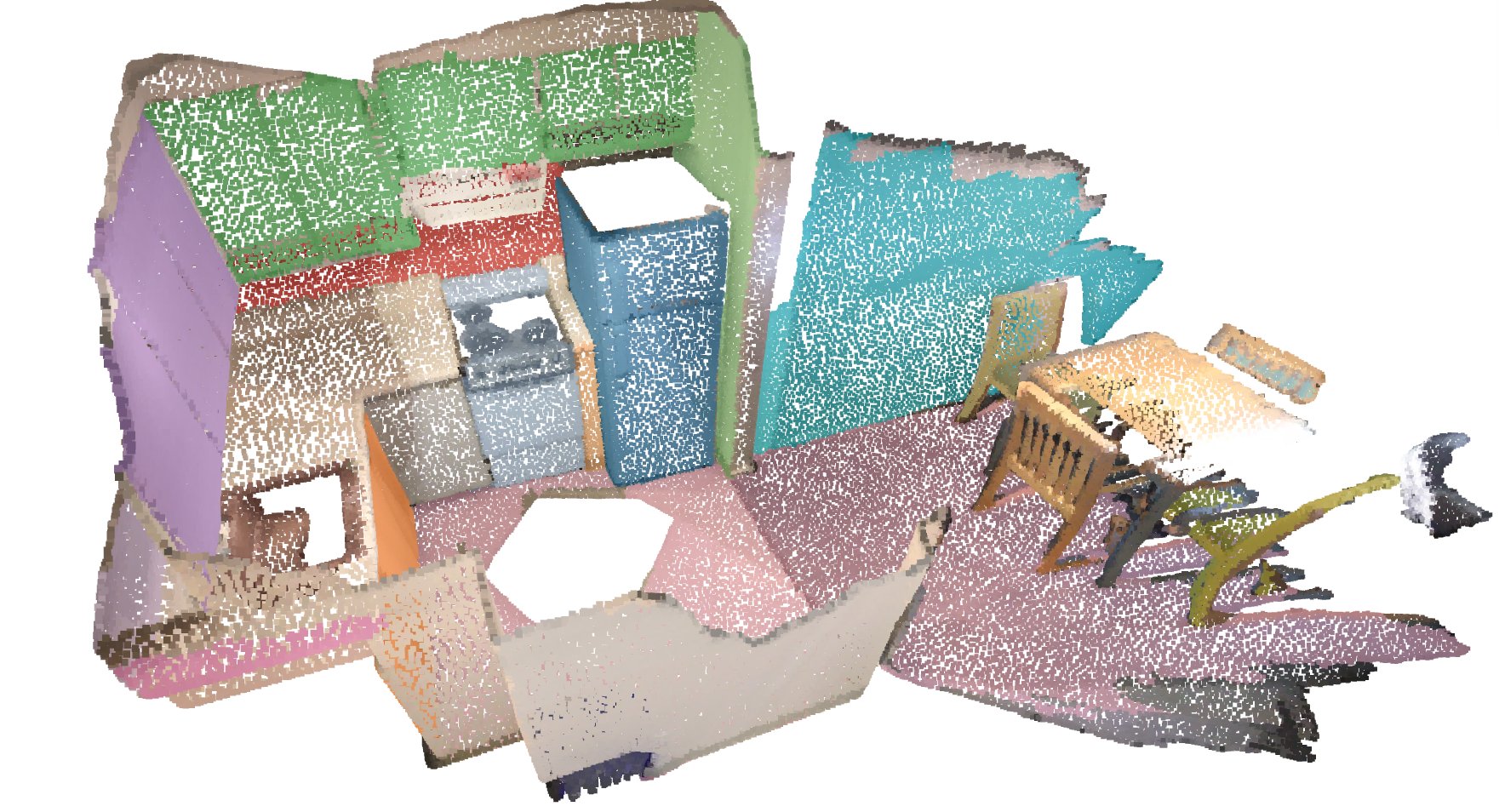} \\
\includegraphics[width=0.32\textwidth]{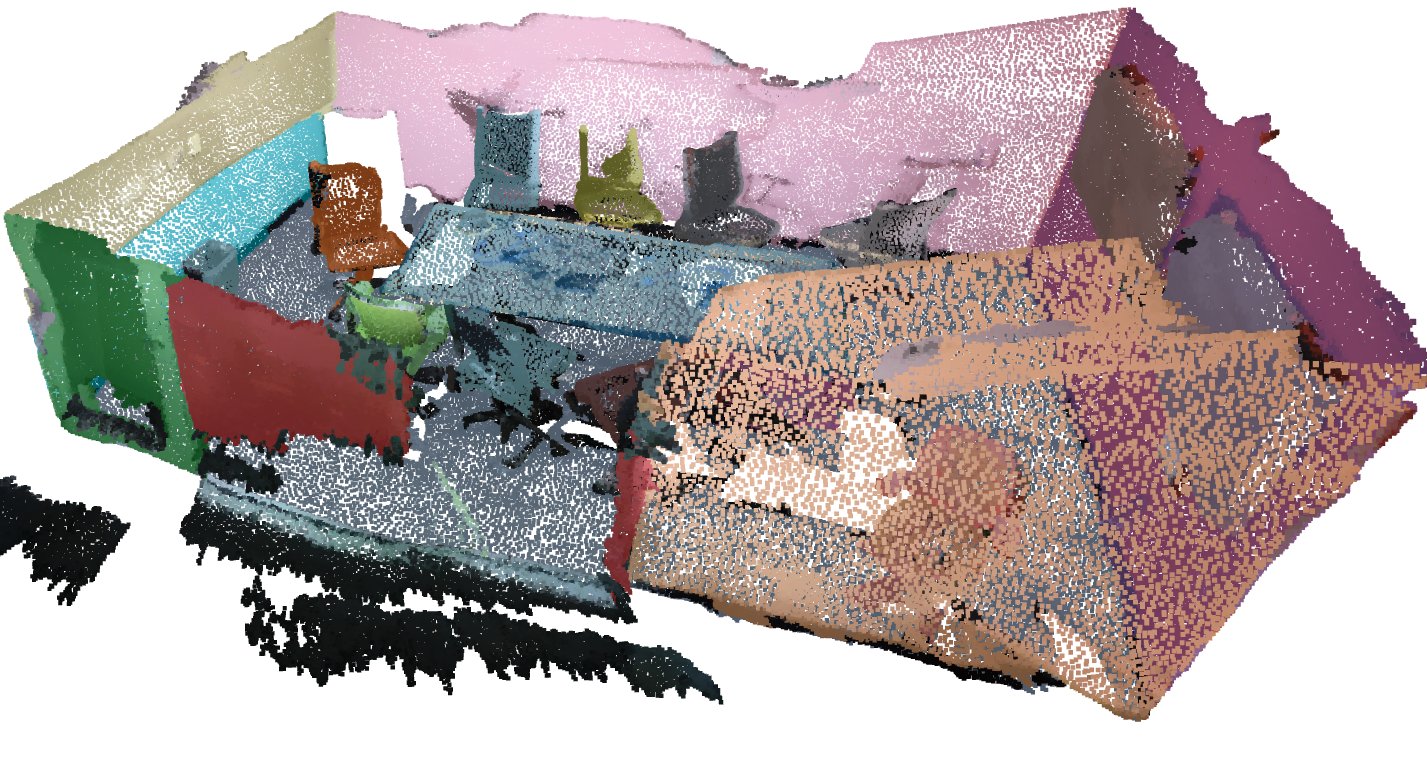} &
\includegraphics[width=0.32\textwidth]{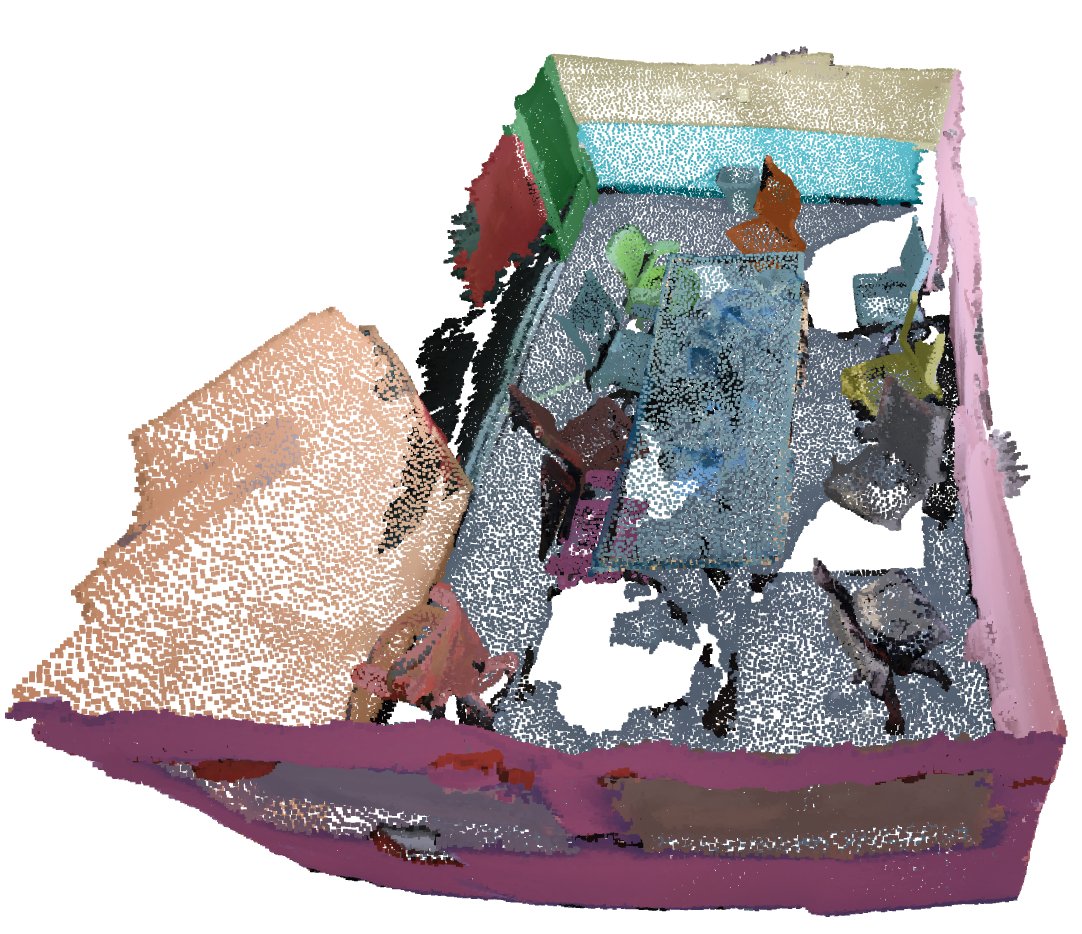} &
\includegraphics[width=0.32\textwidth]{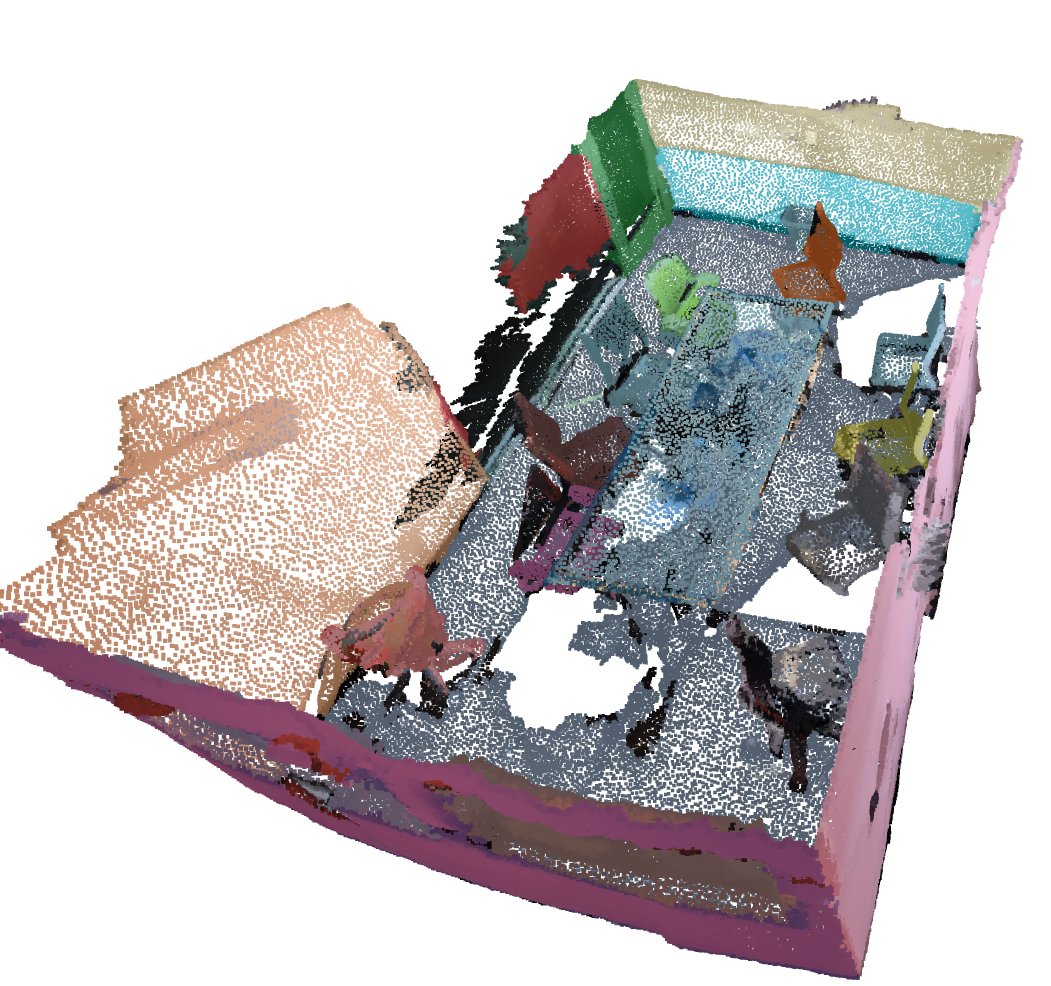} \\
\end{tabular}
\caption{\textbf{Class-agnostic 3D instance segmentation} obtained by projecting 3AM tracks into 3D.}
\label{fig:visual_class}
\end{figure}

\cref{fig:app_vis1,fig:app_vis2,fig:app_vis3,fig:app_vis4,fig:app_vis5,fig:app_vis6,fig:app_vis7,fig:app_vis8} show more video object tracking results on ScanNet++, and \cref{fig:visual_class} shows class-agnostic 3D instance segmentation results.

